\documentclass{eajam}
\renewcommand\thisnumber{x}
\renewcommand\thisyear {200x}
\renewcommand\thismonth{xxx}
\renewcommand\thisvolume{xx}

\usepackage{charter}
\usepackage[charter]{mathdesign}

\usepackage{bm}

\usepackage{soul,color,xcolor}
\usepackage{multirow}

\begin{document}

\markboth{X.Xu, T.Du, W.Kong, B.Shan, Z.Huang and Y.Li}{Convergence of IGD}
\title{Convergence of Implicit Gradient Descent for Training Two-Layer Physics-Informed Neural Networks}


\author[Xianliang Xu, Ting Du, Wang Kong, Bin Shan, Zhongyi Huang and Ye Li]{Xianliang Xu\affil{1}, Ting Du\affil{1}, Wang Kong\affil{2}, Bin Shan\affil{2}, Zhongyi Huang\affil{1} and Ye Li\affil{2}\comma\corrauth}
\address{\affilnum{1}\  Tsinghua University, Beijing, China. \\
	\affilnum{2}\ Nanjing University of Aeronautics and Astronautics, Nanjing, China.}
%
%
\emails{ {\tt xuxl19@mails.tsinghua.edu.cn} (X. Xu), {dt20@mails.tsinghua.edu.cn} (T. Du), {wkong@nuaa.edu.cn} (W. Kong), {bin.shan@nuaa.edu.cn} (B. Shan), {zhongyih@mail.tsinghua.edu.cn} (Z. Huang), {yeli20@nuaa.edu.cn} (Y. Li)}
%
\begin{abstract}
The optimization algorithms are crucial in training physics-informed neural networks (PINNs), as unsuitable methods may lead to poor solutions. Compared to the common gradient descent (GD) algorithm, implicit gradient descent (IGD) outperforms it in handling certain multi-scale problems. In this paper, we provide convergence analysis for the IGD in training over-parameterized two-layer PINNs. We first derive the training dynamics of IGD in training two-layer PINNs. Then, over-parameterization allows us to prove that the randomly initialized IGD converges to a globally optimal solution at a linear convergence rate. Moreover, due to the distinct training dynamics of IGD compared to GD, the learning rate can be selected independently of the sample size and the least eigenvalue of the Gram matrix. Additionally, the novel approach used in our convergence analysis imposes a milder requirement on the network width. Finally, empirical results validate our theoretical findings.
\end{abstract}



\maketitle


\section{Introduction}
Partial differential equations (PDEs) play an important role in modeling phenomena in various fields, such as biology, physics, and engineering. However, solving high-dimensional PDEs numerically is a long-standing challenge in scientific computing due to the curse of dimensionality while utilizing the classic methods, like finite difference, finite volume and finite element methods. In recent years, neural networks have achieved many milestone breakthroughs in the fields of computer vision \cite{10}, natural language processing \cite{11}, and reinforcement learning \cite{12}. The flexibility, scalability and expressive power of neural networks motivate researchers to apply them in the field of scientific computing. Therefore, various methods involving neural networks have been proposed to solve PDEs numerically \cite{6,26,27,28}. The most representative approach along these lines is Physics-Informed Neural Networks (PINNs), which is a flexible and popular framework that incorporates physical knowledge into the architecture of neural networks to solve forward and inverse problems of PDEs.

Specifically, in PINNs, for PDE in the open bounded domain $\Omega\subset \mathbb{R}^d$ with following form:
\begin{equation}
	\left\{
	\begin{aligned}
		\mathcal{F}(u;\theta)(x)&=0, \  x \in \Omega,\\
		\mathcal{B}(u;\theta)(x)&=0, \ x \in \partial \Omega,
	\end{aligned}
	\right.
\end{equation}
where $\theta$ is the parameter of the PDE that could be either a vector or a vector-valued function, $\mathcal{F}$ and $\mathcal{B}$ are differential operators on $\Omega$ and $\partial \Omega$ respectively. Then, one can construct a neural network $u_{w}(x)$ and train it with the following loss function
\begin{equation}
	\int_{\Omega}\left(\mathcal{F}(u_{w};\theta) (x)\right)^2dx+\lambda \int_{\partial \Omega}\left(\mathcal{B}(u_{w};\theta)(y)\right)^2dy,	
\end{equation}
where $\lambda$ is a hyperparameter to balance the two terms from interior and boundary. However, the computation of the integrals can be prohibitively expensive in high dimensions, if using traditional quadrature methods. To circumvent the curse of dimensionality, we can  employ Monte-Carlo sampling to approximate them using a set of collocation points, which yields the following empirical loss function
\begin{equation}
	\frac{|\Omega|}{n_1} \sum\limits_{i=1}^{n_1}(\mathcal{F}(u_{w};\theta)(x_i))^2 +\lambda\frac{|\partial \Omega|}{n_2} \sum\limits_{j=1}^{n_2}(\mathcal{B}(u_{w};\theta)(y_j))^2,	
\end{equation}
where $|\Omega|$ and $|\partial \Omega|$ are the measures of $\Omega$ and $\partial \Omega$ respectively, $\{x_i\}_{i=1}^{n_1}$ and $\{y_j\}_{j=1}^{n_2}$ are i.i.d. random samples according to the uniform distribution $U(\Omega)$ on $\Omega$ and $U(\partial \Omega)$ on $\partial \Omega$, respectively. Thanks to the implementation of automatic differentiation, the empirical loss function (1.3) is computable and can be effectively trained using first-order optimization methods such as SGD, as well as more sophisticated second-order optimizers like L-BFGS.

In fact, the idea of using neural networks to solve PDEs can date back to the last century \cite{25}. Due to the rapid development of  computational power and algorithms, much attention has been drawn to revisiting the idea and various methods involving neural networks to solve PDEs have been proposed. As the most popular approach, PINNs have been applied successfully in various domains, including computational physics \cite{6, 13, 17}, inverse problem \cite{14,15,16} and quantitative finance \cite{18,19,20}. Compared to the classical numerical methods, PINNs method is mesh-less that can be adapted for solving high-dimensional PDEs, like Hamilton-Jacobi-Bellman equation \cite{21}.

There is also a series of works \cite{2,3,4,29,30} trying to understand the optimization of neural networks based on the idea of neural tangent kernel (NTK) \cite{1}. As shown in \cite{1}, the NTK tends to be invariant throughout the gradient flow process as the width of the neural network approaches infinity. As for the neural network with finite width, it requires a more refined analysis to explore the convergence of gradient descent. A breakthrough result is due to \cite{2}, which shows under over-parameterization, gradient descent converges to a globally optimal solution at a linear convergence rate for the $L^2$ regression problems. The crucial insight in \cite{2} is that over-parameterization, in conjunction with random initialization, ensures that every weight does not deviate significantly from its initialization throughout the training process. Later, \cite{9} generalizes the convergence analysis of \cite{2} to two-layer PINNs, showing that gradient descent finds the global optima of
the empirical loss of PINNs under over-parameterization. Compared to gradient descent, implicit gradient descent works better in certain multi-scale problems, which has been shown in \cite{23} numerically. While the work \cite{23} has offered a convergence analysis for the IGD in the context of $L^2$ regression problems, the convergence of IGD for PINNs remains an open problem. In this paper, we aim to fill this gap.

\subsection{Contributions}

\begin{itemize}
	\item We derive the training dynamics of IGD in training two-layer PINNs, which is fundamentally different from GD. This serves as the foundation for our subsequent convergence analysis, and such an approach can also be extended to other optimization algorithms.
	
	\item For two-layer over-parameterized PINNs, we demonstrate that randomly initialized IGD converges to a globally optimal solution at a linear convergence rate. In contrast to the GD, which requires the learning rate $\eta=\mathcal{O}(\lambda_0)$ (where $\lambda_0$ is the smallest eigenvalue of the Gram matrix, as formally defined in Lemma 3.2), the IGD possesses a different training dynamic, leading to that the learning rate of IGD can be chosen independent of the sample size and the least eigenvalue of the Gram matrix. Additionally, when applying our method that is used for PINNs to $L^2$ regression problems, we can obtain an improvement in the learning rate that is similar to the one previously mentioned, which is required to be $\mathcal{O}(\lambda_0/n^2)$ in \cite{23} (where $n$ is the number of training samples).
	
	\item By utilizing the concentration inequality for the sum of independent random variables that are sub-Weibull of order $\alpha>0$ (see the definition in the notation of section 1.3), we can deduce that a width $m=\Omega(poly(\log(n/\delta)))$ is sufficient for the convergence of IGD. This is in contrast to the results in \cite{2,3,9}, where both gradient descent for $L^2$ regression problems and PINNs require a width $m$ that satisfies $m=\Omega(poly(n,1/\delta))$.
	
	\item We validate our conclusions through detailed numerical experiments, demonstrating the efficiency of IGD.
	
\end{itemize}

\subsection{Related Work}

\textbf{Gradient Descent:} It has been shown in \cite{1} that training multi-layer fully-connected neural networks with a smooth activation function via gradient descent is equivalent to performing a certain kernel method as the width of every layer goes to infinity. Inspired by this celebrated result, several works have demonstrated that gradient-based methods like gradient descent can achieve zero training error despite that the objective function is non-convex \cite{1,2,3,4}. For the $L^2$ regression problems, \cite{2} showed that for randomly initialized and over-parameterized two-layer ReLU neural networks, gradient descent converges to a globally optimal solution at a linear convergence rate under the setting that no two inputs are parallel. And similar results have been established in \cite{3} for deep over-parameterized neural networks with smooth activation functions. Both rely on the observation that over-parameterization makes weights close to their initializations for all iterations, which yields the linear convergence. Based on a similar approach to those in \cite{2} and \cite{3}, \cite{23} conducted the convergence analysis of IGD for $L^2$ regression problems, requiring that the learning rate $\eta$ be $\mathcal{O}(\lambda_0/n^2)$ to achieve the convergence. However, the approach cannot be generalized to PINNs due to the presence of derivatives in the loss functions of PINNs.

\textbf{PINNs:} Various optimization methods are available for PINNs, like L-BFGS \cite{6}, natural gradient descent \cite{7}, but none have been accompanied by convergence analysis. \cite{8} proved the global linear convergence of gradient descent in training PINNs for second-order linear PDEs, but only focused on the gradient flow and some assumptions are not verified. The work most relevant to ours is \cite{9}, which derived the rigorous convergence analysis of gradient descent for two-layer PINNs with $\text{ReLU}^3$ activation functions. Compared to \cite{9} for GD, IGD has a completely different training dynamic, which requires a new processing method and yields a free choice for the learning rate. Moreover, the required width of the two-layer neural network in \cite{9} is of order $poly(n,1/\delta)$, which is much worse than our result.

\subsection{Notation} For $x \in \mathbb{R}^d$, we denote by $\|x\|_2 $ the Euclidean norm of $x$. Let $g : [0,\infty)\rightarrow [0,\infty)$ be a non-decreasing function with $g(0)=0$. The $g$-Orlicz norm of a real-valued random variable $X$ is given by
\[\|X\|_{g}:=\inf\left\{t>0: \mathbb{E}\left[g\left(\frac{|X|}{t}\right)\right]\leq 1\right\}.\]
A random variable $X$ is said to be sub-Weibull of order $\alpha>0$, denoted as sub-Weill($\alpha$), if $\|X\|_{\psi_{\alpha}}<\infty$, where
\[\psi_{\alpha}(x):=e^{x^{\alpha}}-1, \quad \text{for} \ x\geq 0. \]
For two positive functions $f_1(n)$ and $f_2(n)$, we use $f_1(n)=\mathcal{O}(f_2(n))$, $f_2(n)=\Omega(f_1(n))$ or $f_1(n)\lesssim f_2(n)$ to represent that $f_1(n)\leq Cf_2(n)$, where $C$ is a universal constant. In this paper, we say that $C$ is a universal constant, meaning that $C$ is a constant that independent of any variables, including sample size, underlying dimension and so on. For $n_1,n_2\in \mathbb{N}$ with $n_1\leq n_2$, we let $[n]=\{1,\cdots, n\}$ and $[n_1,n_2]=\{n_1,\cdots,n_2 \}$. Given a set $S$, we use $Unif\{S\}$ to denote the uniform distribution over $S$.

\section{Preliminaries}
In this paper, in order to compare with the convergence analysis of gradient descent in \cite{9}, we focus on the PDE with the same form:
\begin{equation}
\left\{
\begin{aligned}
&\frac{\partial u}{\partial x_0}(\bm{x})-\sum\limits_{i=1}^d \frac{\partial^2 u}{\partial x_i^2}(\bm{x})=f(\bm{x}), \, \bm{x}\in (0,T)\times \Omega, \\
&u(\bm{x})=g(\bm{x}), \ \bm{x} \in \{0\}\times \Omega \cup [0,T]\times \partial \Omega,
\end{aligned}
\right.
\end{equation}
where $\bm{x}=(x_0,x_1,\cdots,x_d)\in \mathbb{R}^{d+1}$ and $x_0 \in [0,T]$ is the time variable. In the following, we assume that $\|\bm{x}\|_2\leq 1$ for $x\in [0,T]\times \bar{\Omega}$ and $f,g$ are bounded continuous functions.

Moreover, we consider a two-layer neural network of the following form.
\begin{equation}
u(\bm{x};\bm{w},\bm{a})=\frac{1}{\sqrt{m}}\sum\limits_{r=1}^m a_r\sigma(\bm{w}_r^T \bm{x}),
\end{equation}
where $\bm{w}=(\bm{w}_1^T,\cdots, \bm{w}_m^T)^T\in \mathbb{R}^{m(d+1)}$, $\bm{a}=(a_1,\cdots, a_m)^T \in \mathbb{R}^m$ and $\bm{w}_r\in \mathbb{R}^{d+1}$ is the weight vector of the first layer, $a_r$ is the output weight and $\sigma(\cdot)$ is the activation function satisfying the following assumption.

\begin{assumption}
There exists a constant $c>0$ such that $\sup_{z\in \mathbb{R}} |\sigma^{(3)}(z)|\leq c$ and for any $z,z^{'}\in \mathbb{R}$,
\begin{equation}
	\left|\sigma^{(k)}(z)-\sigma^{(k)}(z^{'})\right|\leq c\left|z-z^{'}\right|,
\end{equation}
where $k\in \{0,1,2,3\}$. Moreover, $\sigma(\cdot)$ is analytic and is not a polynomial function.
We also assume that for any positive integer $n\geq 2$, $\lim\limits_{x\to +\infty} \sigma^{(n)}(x)/\phi(x)=c_n\neq 0$, where the function $\phi(\cdot)$ satisfies that  \\
\begin{equation*}
	\lim\limits_{x\to +\infty}\phi(x) = 0, \lim\limits_{x\to +\infty}x\cdot\frac{\phi(bx)}{\phi(x)} = 0
\end{equation*} 
holds for any constant $b>1$.
\end{assumption}

\begin{remark}
Assumption 2.1 holds true for a variety of commonly used activation functions, including softplus function $\sigma(z)=\log(1+e^z)$, logistic function $\sigma(z)=1/(1+e^{-z})$, hyperbolic tangent function $\sigma(z)=(e^z-e^{-z})/(e^z+e^{-z})$, among others. Similar assumptions were made in \cite{3} to demonstrate the stability of the training process for deep neural networks and to ensure the positive-definiteness of the Gram matrices. Smooth activation functions are favored in the applications of neural networks for PDE-related problems, as evidenced by various studies \cite{6,16,21,33}. Furthermore, neural networks with smooth activation functions have been shown to possess powerful approximation abilities, as demonstrated in \cite{34,35}. For the sake of consistency and brevity, throughout this paper, we treat all parameters associated with these activation functions as universal constants.
\end{remark}

In the framework of PINNs, we focus on the empirical risk minimization problem. Given training samples $\{\bm{x}_p\}_{p=1}^{n_1}$ and $\{\bm{y}_j\}_{j=1}^{n_2}$ that are from interior and boundary respectively, we aim to minimize the empirical loss function as follows.
\begin{equation}
\begin{aligned}
&L(\bm{w},\bm{a}):= \\
&\sum\limits_{p=1}^{n_1}\frac{1}{2n_1}\left( \frac{\partial u}{\partial x_0}(\bm{x}_p;\bm{w},\bm{a})-\sum\limits_{i=1}^d \frac{\partial^2 u}{\partial x_i^2}(\bm{x}_p;\bm{w},\bm{a})-f(\bm{x}_p)\right)^2+\sum\limits_{j=1}^{n_2}\frac{1}{2n_2} \left(u(\bm{y}_j;\bm{w},\bm{a})-g(\bm{y}_j)\right)^2,
\end{aligned}
\end{equation}
where we omit the hyperparameter $\lambda$ for brevity, and the convergence analysis can easily extend to the case with $\lambda$.

The gradient descent updates the weights by the following formulations:
\begin{equation}
\begin{aligned}
\bm{w}_r(k+1)&=\bm{w}_r(k)-\eta \frac{\partial L(\bm{w}(k), \bm{a}(k))  }{\partial \bm{w}_r}, \\
a_r(k+1)&=a_r(k)-\eta \frac{\partial L(\bm{w}(k), \bm{a}(k))  }{\partial a_r}
\end{aligned}
\end{equation}
for all $r\in [m]$ and $k\in \mathbb{N}$, where $\eta>0$ is the learning rate.

For a general loss function $L(\theta):\mathbb{R}^m\rightarrow \mathbb{R}$ with a weight vector $\theta \in \mathbb{R}^m$, gradient descent (GD) updates the weights by
\[\theta_{k+1}=\theta_k-\eta \nabla L(\theta_k),\]
where $\eta$ is the learning rate and $k\in \mathbb{N}$. Despite the efficiency of GD in various tasks, including optimizing neural networks, GD may suffer from numerical instability due to the selection of certain key parameters, such as the learning rate. In contrast, implicit gradient descent (IGD) may mitigate numerical instability, with an updating rule of
\[\theta_{k+1}=\theta_k-\eta \nabla L(\theta_{k+1}).\]

Here, we borrow an example from \cite{23} to illustrate the instability of GD and the stability of IGD. For loss function 
\[L(\theta_1,\theta_2)=\frac{K_1}{2} (\theta_1-\theta_1^{*})^2+\frac{K_2}{2} (\theta_2-\theta_2^{*})^2,\]
where $\theta_1, \theta_2$ are parameters to be optimized, $\theta_1^{*}, \theta_2^{*}, K_1, K_2$ (with $K_1>0, K_2>0$) are fixed model parameters.

When updating $\theta_1$ and $\theta_2$ by GD with learning rate $\eta$, a direct computation shows that
\[D:=\frac{L(\theta_1^{k+1}, \theta_2^{k+1})}{L(\theta_1^{k}, \theta_2^{k})}\leq \max\left\{(1-\eta K_1)^2, (1-\eta K_2)^2\right\},\]
where $D$ is the loss decay rate, reflecting the speed of convergence. From this, it is clear that we must set $\max\left\{(1-\eta K_1)^2, (1-\eta K_2)^2\right\}\leq 1$, i.e., $\eta \leq \frac{2}{\max\{K_1,K_2\}}$, to ensure the convergence of GD. However, when $K_1$ and $K_2$ differ significantly in magnitudes, the convergence of GD can become slow. For example, if $K_1=10^{-4}$ and $K_2=10^4$, then this requires that $\eta\leq 10^{-4}$. Although this ensures the convergence of GD, it leads to a very slow convergence rate, as shown by
\[D=\frac{L(\theta_1^{k+1}, \theta_2^{k+1})}{L(\theta_1^{k}, \theta_2^{k})}\leq (1-10^{-8})^2.\]

When applying IGD to optimize $L(\theta_1,\theta_2)$, we have
\[D=\frac{L(\theta_1^{k+1}, \theta_2^{k+1})}{L(\theta_1^{k}, \theta_2^{k})}\leq \max\left\{\frac{1}{(1+\eta K_1)^2}, \frac{1}{(1+\eta K_2)^2}\right\}.\]
Thus, regardless of our choice of $\eta$, it always results in $D<1$. Furthermore, we can take a sufficiently large $\eta$ such that $D<\frac{1}{2}$, leading to a faster convergence rate compared to GD.

In addressing our problem, we employ the IGD algorithm, which provides the following updating procedure for the weight vectors.
\begin{equation}
\begin{aligned}
\bm{w}_r(k+1)&=\bm{w}_r(k)-\eta \frac{\partial L(\bm{w}(k+1), \bm{a}(k+1))  }{\partial \bm{w}_r}, \\
a_r(k+1)&=a_r(k)-\eta \frac{\partial L(\bm{w}(k+1), \bm{a}(k+1))  }{\partial a_r}.
\end{aligned}
\end{equation}
In this paper, we consider the initialization that is same as that used in the $L^2$ regression problem [2]:
\begin{equation}
\bm{w}_r(0)\sim \mathcal{N}(\bm{0},\bm{I}_{d+1}), \, a_r(0)\sim Unif(\{-1,1\}).
\end{equation}
For simplicity, we fix the second layer and apply IGD to optimize the first layer, i.e., only optimize $\bm{w}$ with
\begin{equation}
	\bm{w}(k+1)=\bm{w}(k)-\eta \nabla L(\bm{w}(k+1)) ,
\end{equation}
where $L(\bm{w}(k+1))$ is an abbreviation of $L(\bm{w}(k+1),\bm{a})$.

Note that the updating rule for $\bm{w}$ is equivalent to that
\begin{equation}
	\bm{w}(k+1)=\mathop{\arg\min}\limits_{\bm{w}} \frac{1}{2} \|\bm{w}-\bm{w}(k)\|_2^2+ \eta L(\bm{w}).
\end{equation}
For the sub-optimization problem (2.9), we apply L-BFGS or Adam to compute $\bm{w}(k+1)$.
Since the sub-optimization problem has good initial point $\bm{w}(k)$, thus are easier for L-BFGS to achieve good convergence properties.
We replace L-BFGS by Adam when the parameters of PINNs are too large for the quasi-Hessian matrix computation.
The detailed optimization process is the same as the Algorithm 1 in \cite{23}.

\section{Main Results}
In this section, we show that initialized IGD converges to the global minimum at a linear rate. Before presenting the results, we first simplify the notations. For the residuals of interior and boundary, we denote them by $s_p(\bm{w})$ and $h_j(\bm{w})$ respectively, i.e.,
\begin{equation}
s_p(\bm{w})=\frac{1}{\sqrt{n_1}} \left( \frac{\partial u}{\partial x_0}(\bm{x}_p;\bm{w})-\sum\limits_{i=1}^d \frac{\partial^2 u}{\partial x_i^2}(\bm{x}_p;\bm{w})-f(\bm{x}_p)\right)
\end{equation}
and
\begin{equation}
h_j(\bm{w})=\frac{1}{\sqrt{n_2}} \left(u(\bm{y}_j;\bm{w})-g(\bm{y}_j)\right).
\end{equation}
Then the empirical loss function (2.4) can be rewritten as
\begin{equation}
L(\bm{w}):=\frac{1}{2}(\|\bm{s}(\bm{w})\|_2^2+ \|\bm{h}(\bm{w})\|_2^2),
\end{equation}
where
\begin{equation}
\bm{s}(\bm{w})=(s_1(\bm{w}), \cdots, s_{n_1}(\bm{w}))^T\in \mathbb{R}^{n_1},	
\end{equation}
and
\begin{equation}
\bm{h}(\bm{w})=(h_1(\bm{w}), \cdots, h_{n_2}(\bm{w}))^T \in \mathbb{R}^{n_2}.
\end{equation}

At this time, we have
\begin{equation}
\begin{aligned}
\frac{\partial L(\bm{w})}{\partial \bm{w}_r}=\sum\limits_{p=1}^{n_1}s_p(\bm{w})\frac{\partial s_p(\bm{w})}{\partial \bm{w}_r}+\sum\limits_{j=1}^{n_2}h_j(\bm{w})\frac{\partial h_j(\bm{w})}{\partial \bm{w}_r},
\end{aligned}
\end{equation}
which can be used to derive the recursive formula in the following lemma for the IGD.

\begin{lemma}
For the implicit gradient descent, we can deduce that for all $k\in \mathbb{N}$,
\begin{equation}
\begin{aligned}
&\begin{bmatrix}
    \bm{s}(\bm{w}(k+1))\\ \bm{h}(\bm{w}(k+1))
\end{bmatrix} =(\bm{I}+\eta \bm{G}(k+1))^{-1}\left( \begin{bmatrix}
    \bm{s}(\bm{w}(k))\\ \bm{h}(\bm{w}(k))
\end{bmatrix} - \begin{bmatrix}
    \bm{I}_1(k+1)\\ \bm{I}_2(k+1)
\end{bmatrix}\right),
\end{aligned}
\end{equation}
where $\bm{G}(t)$ is the Gram matrix in the $t$-th ($t\in\mathbb{N}$) iteration, defined as
\begin{equation}
\bm{G}(t)=\bm{D}(t)^T\bm{D}(t)
\end{equation}
with definition of $\bm{D}(t)$ as
\begin{equation}
\left[ \frac{\partial s_1(\bm{w}(t))}{\partial \bm{w} },\cdots, \frac{\partial s_{n_1}(\bm{w}(t))}{\partial \bm{w} }, \frac{\partial h_1(\bm{w}(t))}{\partial \bm{w} }, \cdots, \frac{\partial h_{n_2}(\bm{w}(t))}{\partial \bm{w} }\right]	
\end{equation}
and $\bm{I}_1(t)$, $\bm{I}_2(t)$ are residual terms, the $p$-th ($1\leq p\leq n_1$) component of $\bm{I}_1(t)$ is defined as
\begin{equation}
\int_{0}^{\eta} \left\langle \nabla L(\bm{w}(t)), \nabla s_p(\bm{w}(t)+\alpha\nabla L(\bm{w}(t)) )- \nabla s_p(\bm{w}(t))\right\rangle d\alpha,	
\end{equation}
the $j$-th ($1\leq j\leq n_2$) component of $\bm{I}_2(t)$ is defined as
\begin{equation}
\int_{0}^{\eta} \left\langle \nabla L(\bm{w}(t)), \nabla h_j(\bm{w}(t)+\alpha\nabla L(\bm{w}(t)) )- \nabla h_j(\bm{w}(t))\right\rangle d\alpha.
\end{equation}
\end{lemma}

Lemma 3.1 shows that the implicit gradient descent has a completely different dynamic from the gradient descent, which admits the following recursive formula.
\begin{equation}
\begin{bmatrix}
    \bm{s}(\bm{w}(k+1))\\ \bm{h}(\bm{w}(k+1))
\end{bmatrix} =(\bm{I} -\eta \bm{G}(k))\begin{bmatrix}
    \bm{s}(\bm{w}(k))\\ \bm{h}(\bm{w}(k))
\end{bmatrix}-\begin{bmatrix}
    \bm{\chi}_1(k)\\ \bm{\chi}_2(k)
\end{bmatrix},
\end{equation}
where $\bm{\chi}_1(k)$ and $\bm{\chi}_2(k)$ are residual terms that are similar to $\bm{I}_1(k+1)$ and $\bm{I}_2(k+1)$. We refer to them as residual terms because we can later prove that for all
$k\in \mathbb{N}$, the following holds (as shown in Lemma 3.9):
\[ \|\bm{I}_1(k+1)\|_2=\mathcal{O}\left(\frac{1}{\sqrt{m}}\right), \quad \|\bm{I}_2(k+1)\|_2=\mathcal{O}\left(\frac{1}{\sqrt{m}}\right)\]
Roughly speaking, this result implies that when $m$ is sufficiently large, the impact of these two terms on the overall result is negligible.

\begin{remark}
The recursive formula (3.7) is crucial in our method for convergence analysis of IGD. As for GD, \cite{9} does not derived the recursive formula (3.12) for GD. The proof strategy in \cite{9} is the same as that in \cite{2}. In brief, it relies on the decomposition
\[\bm{s}(\bm{w}(k+1))=\bm{s}(\bm{w}(k))+(\bm{s}(\bm{w}(k+1))-\bm{s}(\bm{w}(k)))\]
and the upper bound for $\| \bm{s}(\bm{w}(k+1))-\bm{s}(\bm{w}(k))\|_2$. This decomposition and upper bound are the reason for the requirement of $\eta$, i.e., $\eta=\mathcal{O}(\lambda_0)$. Note that $\lambda_0$ may be extremely small in practice due to its dependence on the sample size, which can make the training process struggle. In contrast, different training dynamic and utilization of the recursive formula (3.7) enable us to demonstrate that the learning rate of IGD is independent of $n_1,n_2$ and $\lambda_0$ and can be chosen freely. Of independent interest, when using the recursive formula (3.12) in the convergence analysis of gradient descent, we may obtain a milder requirement for the learning rate. In fact, the requirement that $\eta \|\bm{G}(k)\|_2\lesssim 1$ for all $k\in \mathbb{N}$ is sufficient for the convergence of GD. Given that $\|\bm{G}(k)-\bm{G}^{\infty}\|_2=\mathcal{O}(1/\sqrt{m})$ holds for all $k\in\mathbb{N}$, it follows that the condition $\eta \|\bm{G}^{\infty}\|_2\lesssim 1$ suffices for the convergence when $m$ is large enough. Furthermore, (3.7) and (3.12) provide insights similar to those shown in the example of preliminaries section. Specifically, in (3.12), we require $\bm{I}-\eta \bm{G}(k)$ to be positive definite for GD, hence $\eta \|\bm{G}^{\infty}\|_2\lesssim 1$. In contrast, in (3.7), $\bm{I}+\eta \bm{G}(k+1)$ is always strictly positive definite, leading to no requirements for $\eta$.
\end{remark}

Similar to the convergence analysis of gradient descent for the $L^2$ regression problems, we first define the Gram matrix $\bm{G}^{\infty} \in \mathbb{R}^{(n_1+n_2)\times (n_1+n_2)}$ as
$\bm{G}^{\infty}=\mathbb{E}_{\bm{w}(0)\sim \mathcal{N}(\bm{0},\bm{I}) }\bm{G}(0)$. From \cite{37}, we know that under mild conditions on the training samples, the Gram matrix is strictly positive definite, as stated in Lemma 3.2.

\begin{lemma}
If no two samples in $\{\bm{x}_p\}_{p=1}^{n_1}\cup \{\bm{y}_j\}_{j=1}^{n_2}$ are parallel, then $\bm{G}^{\infty}$ is strictly positive definite for activation functions that satisfy Assumption 2.1, i.e., $\lambda_0:=\lambda_{min}(\bm{G}^{\infty})>0$.
\end{lemma}

\begin{remark}
Due to the fact that two points in the high-dimensional space tend to be parallel, we can consider the neural network with bias. Specifically, we can replace the sample $\bm{x}$ with $(\bm{x}^T,1)^T$, then the non-parallel condition in Lemma 3.2 is equivalent to that no two samples are equal, which is natural to hold.
\end{remark}

Under over-parameterization, we can deduce that at random initialization, $\bm{G}(0)$ is sufficiently close to $\bm{G}^{\infty}$, ensuring that $\bm{G}(0)$ has a lower bounded least eigenvalue with high probability. In the context of the $L^2$ regression problem, since all related random variables are bounded, the Hoeffding's inequality or Bernstein's inequality suffices. However, due to the involvement of derivatives in the framework of PINNs, the random variables are not bounded anymore. Fortunately, we can demonstrate that they are sub-Weibull of order $\alpha>0$, and thus the concentration inequality for the sum of sub-Weibull variables is available. We provide the detailed proof in the appendix.

\begin{lemma}
If $m=\Omega\left(\frac{d^4}{\lambda_0^2}\log\left(\frac{n_1+n_2}{\delta}\right)\right)$, we have that with probability at least $1-\delta$, $\|\bm{G}(0)-\bm{G}^{\infty}\|_2\leq \frac{\lambda_0}{4}$ and $\lambda_{min}(\bm{G}(0))\geq \frac{3}{4}\lambda_0$.
\end{lemma}

Over-parameterization not only ensures that $\bm{G}(0)$ is strictly positive definite in high probability, but also that $\bm{G}(0)$ is stable under small perturbations. Formally, as shown in the following lemma, if $\bm{w}_r$ is close to its random initialization $\bm{w}_r(0)$ for all $r\in[m]$, the induced Gram matrix $\bm{G}(\bm{w})$ is also close to $\bm{G}(0)$. Thus, if the hidden weight vectors do not go far away from their initializations during the training process, the Gram matrix in each iteration stays close to $\bm{G}(0)$.

\begin{lemma}
Let $R\in (0,1]$, if $\bm{w}_1(0),\cdots,\bm{w}_m(0)$ are i.i.d. generated from $\mathcal{N}(\bm{0},\bm{I}_{d+1})$, then with probability at least $1-\delta$, the following holds. For any set of weight vectors $\bm{w}_1,\cdots,\bm{w}_m \in \mathbb{R}^{d+1}$ that satisfy that for any $r\in [m]$, $\|\bm{w}_r-\bm{w}_r(0)\|_2\leq R$, then the induced Gram matrix $\bm{G}(\bm{w})\in \mathbb{R}^{(n_1+n_2)\times (n_1+n_2)}$ satisfies
\begin{equation}
\begin{aligned}
    &\|\bm{G}(\bm{w})-\bm{G}(0)\|_2 \leq CR\max\left\{d^2,\frac{d^2}{\sqrt{m}}\sqrt{\log\left(\frac{1}{\delta}\right)}, \frac{d^2}{m}\left(\log\left(\frac{1}{\delta}\right)\right)^2 \right\},
\end{aligned}
\end{equation}
where $C$ is a universal constant.
\end{lemma}

With the preparations above, we next prove the convergence of IGD by induction. In the proof, we first show that when $m$ is large enough, the weight vector $\bm{w}_r$ is close to its random initialization $\bm{w}_r(0)$ for all $r\in [m]$ and all iterations. Then Lemma 3.4 implies that the corresponding Gram matrix has a lower bounded least eigenvalue. By bounding the residual terms in the recursive formula of Lemma 3.1, we arrive at our conclusion. Formally, our induction hypothesis concerns the convergence of the empirical loss and the boundedness of the weight vectors.

\textbf{Condition 1:} At the $k$-th iteration, we have
$\|\bm{w}_r(k)\|_2 \leq B$ for all $r\in [m]$ and
\begin{equation}
	L(k)\leq \left(1+\frac{\eta\lambda_0}{2}\right)^{-k}L(0),
\end{equation}
where $L(k)$ is an abbreviation of $L(\bm{w}(k))$ and $B>1$, is a constant defined in (3.17).

The assumption of the boundedness of $\|\bm{w}_r(k)\|_2$ is necessary, given the presence of derivatives in the empirical loss function. Observing that $\|\bm{w}_r(k)\|_2 \leq \|\bm{w}_r(k)-\bm{w}_r(0)\|_2+\|\bm{w}_r(0)\|_2$, the boundedness assumption holds directly from the facts that $\bm{w}_r(k)$ is close to $\bm{w}_r(0)$ under over-parameterization and $\bm{w}_r(0)$ is bounded with high probability for all $r\in[m]$. The boundedness of the weights and initial value is also due to the utilization of concentration inequality for sub-Weibull random variables, and we put the detailed proofs in Appendix.

\begin{lemma}
With probability at least $1-\delta$,
\begin{equation}
\|\bm{w}_r(0)\|_2^2 \leq C\left(d+\sqrt{d\log\left(\frac{m}{\delta}\right)}+\log\left(\frac{m}{\delta}\right) \right)
\end{equation}
holds for all $r\in [m]$ and $C$ is a universal constant.
\end{lemma}

\begin{lemma}
With probability at least $1-\delta$, we have
\begin{equation}
    L(0)\leq C \left(d^2\log\left(\frac{n_1+n_2}{\delta}\right)+\frac{d^2}{m} \left(\log\left(\frac{n_1+n_2}{\delta}\right)\right)^2\right),	
\end{equation}
where $C$ is a universal constant.
\end{lemma}

From Lemma 3.5 and the fact that $\|\bm{w}_r(k)\|_2 \leq \|\bm{w}_r(k)-\bm{w}_r(0)\|_2+\|\bm{w}_r(0)\|_2$, we can let
\begin{equation}
B=\sqrt{ C\left(d+\sqrt{d\log\left(\frac{m}{\delta}\right)}+\log\left(\frac{m}{\delta}\right) \right) }+1,
\end{equation}
where $C$ is the constant in (3.15).

With above preparations, we can now verify that all weight vectors are close to their initializations throughout the training process.

\begin{lemma}
If Condition 1 holds for $t=0,\cdots,k$, then we have that for every $r\in [m]$
\begin{equation}
\begin{aligned}
    &\|\bm{w}_r(k+1)-\bm{w}_r(0)\|_2\leq \frac{C\eta}{\sqrt{m}} (\eta L(0)+B^2)\sqrt{L(0)} +\frac{C B^2}{\sqrt{m}\lambda_0} \sqrt{L(0)}.
\end{aligned}
\end{equation}	
\end{lemma}

Lemma 3.7 shows that for sufficiently large $m$, $\bm{w}_r(k+1)$ is close enough to $\bm{w}_r(0)$ for all $r\in [m]$. Therefore, combining the stability property of the Gram matrix, i.e., Lemma 3.4, yields the following lemma. 

\begin{lemma}
When $m$ satisfies that
\begin{equation}
\begin{aligned}
    m=&\Omega\left(\frac{d^8}{\lambda_0^2}\log\left(\frac{n_1+n_2}{\delta}\right)\max\left\{\eta^4 d^2 \log\left(\frac{n_1+n_2}{\delta}\right), \frac{1}{\lambda_0^2} \left(\log\left(\frac{m}{\delta}\right)\right)^2 \right\}  \right)
\end{aligned}
\end{equation}
we have $\lambda_{min}(\bm{G}(k+1))\geq \frac{\lambda_0}{2}$ under the events in Lemma 3.4, Lemma 3.5 and Lemma 3.6, where $C$ is a universal constant. Moreover, $\|\bm{w}_r(k+1)\|_2\leq B$ holds for all $r\in [m]$, where $B$ is the constant defined in (3.17).
\end{lemma}

The induction is not yet complete, it remains only to verify that (3.14) in Condition 1 also holds for $k+1$, i.e.,
\begin{equation}
	L(k+1)\leq \left(1+\frac{\eta\lambda_0}{2}\right)^{-(k+1)}L(0),	
\end{equation}
which can be achieved by bounding the residual terms in the recursive formula (3.7).

\begin{lemma}
Under the setting of Lemma 3.7, we have
\begin{equation}
\|\bm{I}_1(k+1)\|_2 \leq \frac{C\eta^2 B^6}{\sqrt{m}}L(k+1)
\end{equation}
and
\begin{equation}
\|\bm{I}_2(k+1)\|_2 \leq \frac{C\eta^2B^4 }{\sqrt{m}}L(k+1),
\end{equation}
where $C$ is a universal constant.
\end{lemma}

Lemma 3.9 shows that $\|\bm{I}_1(k+1)\|_2=\mathcal{O}(1/\sqrt{m})$ and $\|\bm{I}_2(k+1)\|_2=\mathcal{O}(1/\sqrt{m})$, which is why we refer to them as residual terms in Lemma 3.1. Thus, when $m$ is large enough, we have
\begin{equation}
\begin{aligned}
\begin{bmatrix} \bm{s}(\bm{w}(k+1)) \\ \bm{h}(\bm{w}(k+1) \end{bmatrix}  \approx (\bm{I}+\eta \bm{G}(k+1))^{-1} \begin{bmatrix} \bm{s}(\bm{w}(k)) \\ \bm{h}(\bm{w}(k) \end{bmatrix}. 
\end{aligned}
\end{equation}
Combining this with $\lambda_{min}(\bm{G}(k+1))\geq \frac{\lambda_0}{2}$, which can be derived from Lemma 3.4 and Lemma 3.7, yields the following convergence result for IGD. From this, we can see that it indeed has no requirements for the learning rate $\eta$. 

\begin{theorem}\label{th-lr}
Under Assumption 2.1 for the activation function and the non-parallel condition in Lemma 3.2 for the training samples, we have that with probability at least $1-\delta$ over the random initialization, the implicit gradient descent algorithm satisfies that for all $k\in\mathbb{N}$
\begin{equation}
L(k)\leq \left(1+\frac{\eta\lambda_0}{2}\right)^{-k}L(0),
\end{equation}
where
\begin{equation}
\begin{aligned}
&m=\Omega\left(\frac{d^8}{\lambda_0^2}\log\left(\frac{n_1+n_2}{\delta}\right)\max\left\{\eta^4 d^2 \log\left(\frac{n_1+n_2}{\delta}\right) \frac{1}{\lambda_0^2} \left(\log\left(\frac{m}{\delta}\right)\right)^2, \frac{\eta^2 \left(\log\left(\frac{m}{\delta}\right)\right)^6\log\left(\frac{n_1+n_2}{\delta}\right) }{\lambda_0^2} \right\}  \right)
\end{aligned}
\end{equation}
and $\eta$ can be chosen arbitrarily.
\end{theorem}

\begin{remark}
Note that the conclusion (3.24) holds under the intersection of events in the Lemma 3.3, Lemma 3.4, Lemma 3.5 and Lemma 3.6, which are independent of the iteration $k$.
\end{remark}

\begin{remark}
We first compare our result (3.24) with the state-of-art result in \cite{32} for the $L^2$ regression problems. Specifically, it requires that $m=\Omega\left(\frac{n^4}{\lambda_0^4} \log \left(\frac{m}{\delta}\right) \log^2 \left(\frac{n}{\delta}\right)\right)$ and $\eta=\mathcal{O}\left(\frac{\lambda_0}{n^2}\right)$ to ensure the convergence of gradient descent. Firstly, (3.25) depends polynomially on the dimension $d$, leading to a strict requirement for $m$ when $d$ is large. This is due to the fact that the loss of PINNs directly involves $\|\bm{w}\|_2$, while the loss of $L^2$ regression problem only depends on $\bm{w}^T \bm{x}$. Actually, we may eliminate the dependence on $d$ by initializing $\bm{w}_r(0)\sim \mathcal{N}(\bm{0},\frac{1}{d+1} \bm{I}_{d+1})$ for all $r\in [m]$. Secondly, (3.25) seems almost independent of $n_1$ and $n_2$. This is because we have normalized the loss function of PINNs. If we let $n_1=n_2=n$ and $\widetilde{\lambda_0}=\widetilde{\bm{G}^{\infty}}$, which is induced by the unnormalized loss function of PINNs, then $\lambda_0=\widetilde{\lambda_0}/n$ and the requirement in (3.25) for $m$ becomes $m=\Omega\left(\frac{n^4}{\widetilde{\lambda_0}^4}\right)$, which is same as that for the regression problems. 
\end{remark}

\begin{remark}
When comparing with the convergence results of GD for PINNs, i.e., Theorem 4.5 in \cite{9}, it is required that $m=\widetilde{\Omega}\left( \frac{(n_1+n_2)^2}{\lambda_0^4 \delta^3} \right)$, where $\widetilde{\Omega}(\cdot)$ denotes that some terms involving $\log(m)$ are omitted. Firstly, \cite{9} does not explicitly show the dependence on the underlying dimension $d$. Secondly, the requirement for $m$ depends polynomially on $n_1,n_2$ and $1/\delta$. In contrast, our result (3.25) is independent of $n_1,n_2$ when logarithmic terms about $n_1,n_2$ are omitted and depends only polynomially on $\log(1/\delta)$. Thirdly, the convergence result in \cite{9} requires that the learning rate $\eta$ be $\mathcal{O}(\lambda_0)$ to achieve the linear convergence, which is a stricter requirement than our own. 
\end{remark}

\section{Computational Results}
\subsection{Poisson Equation}
First, we consider the Poisson equation on the domain $\Omega=[0,1]\times[0,1]$,
\begin{equation}
\left.\left\{
\begin{array}
{c}-\frac{\partial^{2} u}{\partial x^{2}}-\frac{\partial^{2} u}{\partial y^{2}}=f(x,y),\quad(x,y)\in \Omega, \\
u(x,y)=0,\quad(x,y)\in \partial \Omega.
\end{array}\right.\right.
\end{equation}
The true solution is chosen as $u(x, y) = \sin(\pi x) \sin(\pi y) + 0.1 \sin(10 \pi x) \sin(10 \pi y)$ with multi-scale features.

\begin{figure}[htbp]
	\centering
	\includegraphics[width=0.8\textwidth]{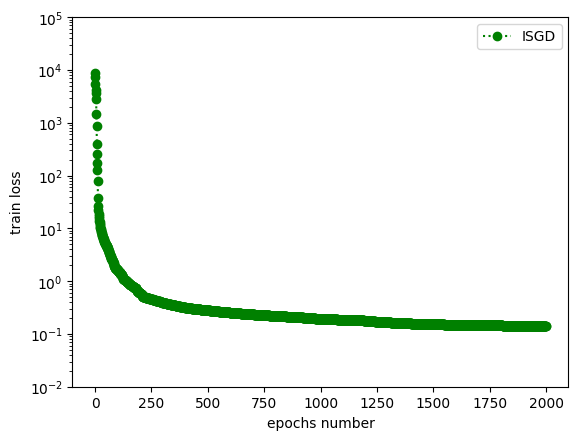}
	\caption{Training loss for the Poisson equation.}
	\label{loss-poisson}
\end{figure}

\begin{figure}[htbp]
	\centering
	\includegraphics[width=\textwidth]{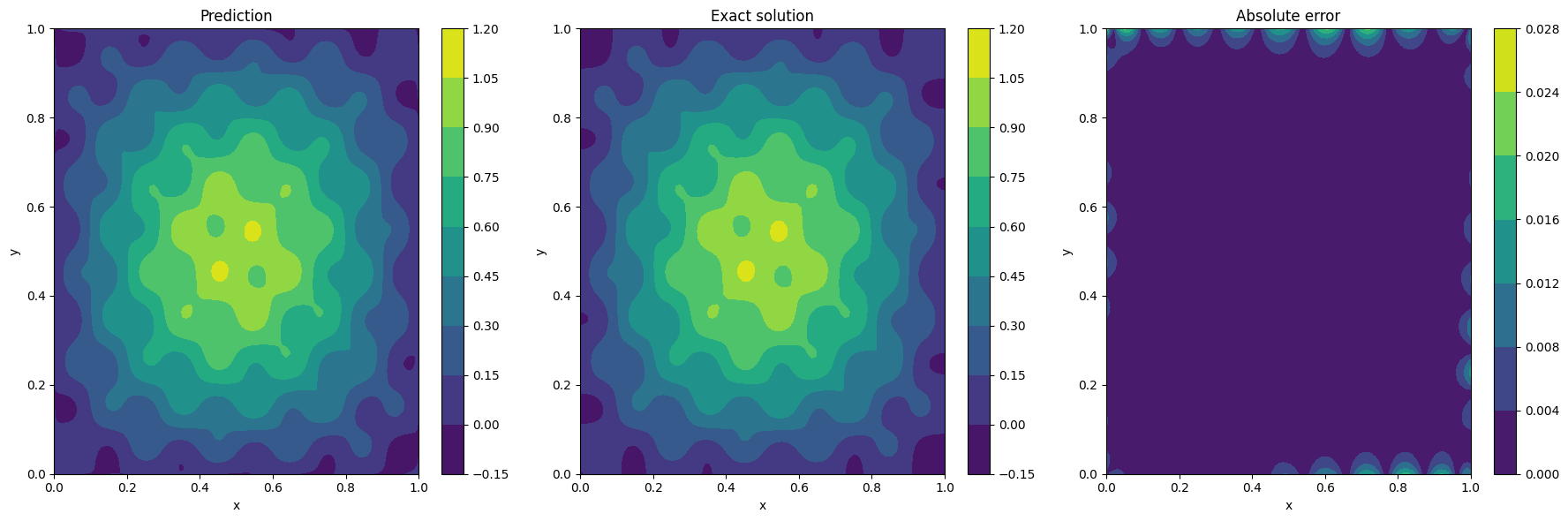}
	\caption{IGD Prediction and Analysis for Poisson Equation.}
	\label{comparison-poisson}
\end{figure}

We sample $N_{b}=400$ random points on the boundary $\partial \Omega$, and $N_{f}=2,000$ random points on the domain $\Omega$. A neural network with $6$ hidden layers, every $128$ units with $tanh$ activations, is applied in all computations. Here we train $2,000$ epochs for the learning rate $lr=0.5$. The relative $L^{2}$ error is $0.5 \%$. In Figure~\ref{loss-poisson}, we can see that our IGD method is capable of achieving an accurate solution with a small number of training epochs, even when utilizing a large learning rate. Figure~\ref{comparison-poisson} illustrates the prediction of the Poisson equation, along with the exact solution and the absolute error between them. It is evident that the prediction aligns closely with the exact solution, demonstrating the superiority of our method.

To better align the theoretical analysis with the experiments, we train shallow neural networks with a single hidden layer of 128 neurons.
The relative $L^{2}$ error is $0.31\%$, show that the IGD algorithm still performs well under the shallow setting, which is consistent with our theoretical analysis.

\subsection{Helmholtz equation}
The Helmholtz equation is one of the fundamental equations of mathematical physics arising in many physical problems, such as vibrating membranes, acoustics, and electromagnetism equations. We solve the 2-D Helmholtz equation on the domain $\Omega=[0,1]\times[0,1]$ given by
\begin{equation}
	\left\{
	\begin{array}
		{l}\frac{\partial^{2} u}{\partial x^{2}}+\frac{\partial^{2} u}{\partial y^{2}}+k^{2} u(x,y)=f(x,y),\quad(x,y)\in \Omega, \\
		u(x,y)=0,\quad(x,y)\in \partial \Omega.
	\end{array}
	\right.
\end{equation}
The exact solution for $k=4$ is $u(x, y) = \sin(\pi x) \sin(4 \pi y)$, and the force term $f(x,y)$ can be directly computed. 

\begin{figure}[htbp]
	\centering
	\includegraphics[width=0.8\textwidth]{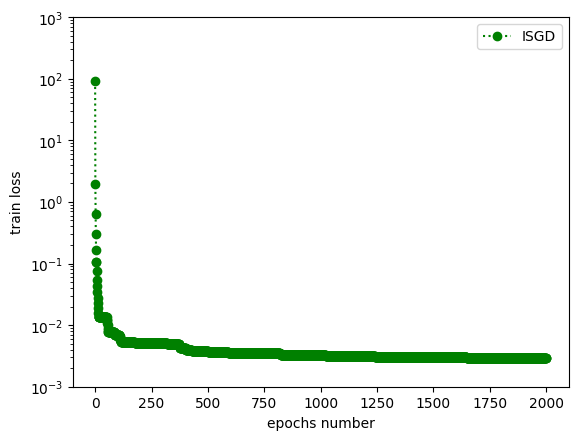}
	\caption{Training loss for the Helmholtz equation.}
	\label{loss-helmholtz}
\end{figure}

\begin{figure}[htbp]
	\centering
	\includegraphics[width=\textwidth]{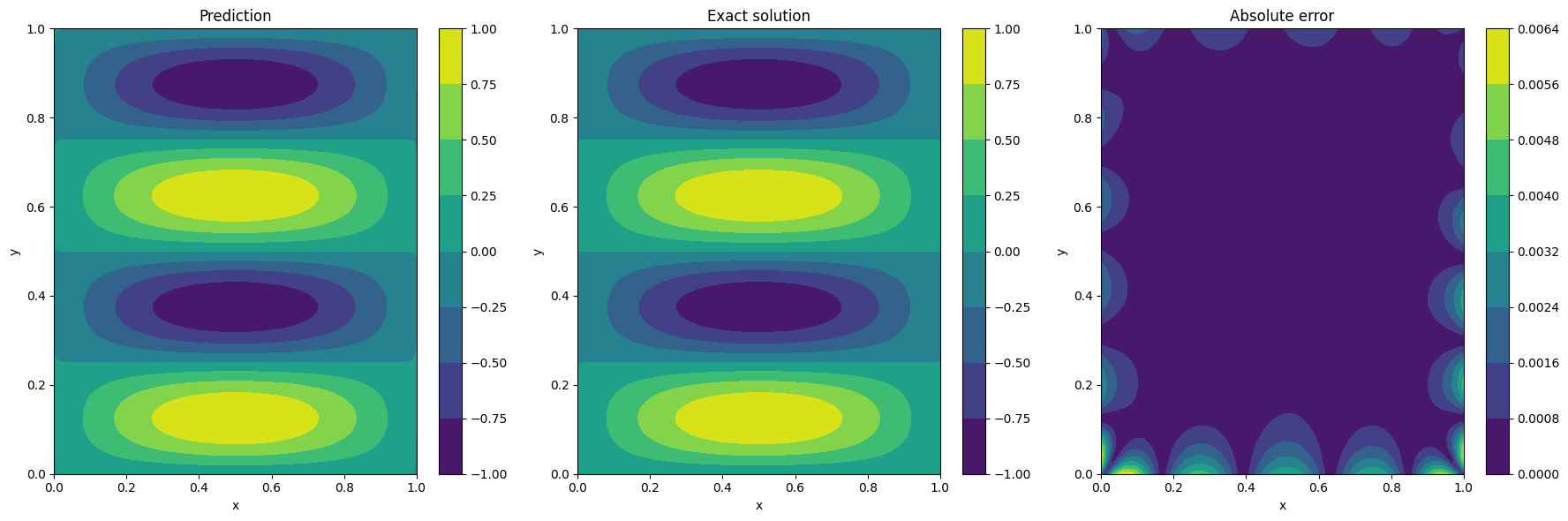}
	\caption{IGD Prediction and Analysis for Helmholtz Equation.}
	\label{comparison-helmholtz}
\end{figure}

We choose $N_{b}=400$ randomly sampled points on the boundary $\partial \Omega$ and $N_{f}=2,000$ randomly sampled points within the domain $\Omega$. A neural network with $6$ hidden layers, each containing $128$ units with $\tanh$ activation functions, is used for all computations. The training process runs for $2,000$ epochs with a learning rate of $lr=0.5$. The relative $L^{2}$ error is $0.17 \%$. As shown in Figure~\ref{loss-helmholtz}, our IGD method achieves high accuracy within a limited number of training epochs, even when using a large learning rate. Figure~\ref{comparison-helmholtz} presents the prediction for the Helmholtz equation, together with the exact solution and the corresponding absolute error. The close agreement between the prediction and the exact solution highlights the effectiveness and robustness of our approach.

To better align the theoretical analysis with the experiments, we train shallow neural networks with a single hidden layer of 128 neurons.
The relative $L^{2}$ error is $0.19\%$, show that the IGD algorithm still performs well under the shallow setting, which is consistent with our theoretical analysis.

\subsection{Allen-Cahn Equation}
We investigated the 2-D Allen-Cahn equation in the domain $\Omega=[0,1]\times[0,1]$, representing non-linear diffusion-reaction processes. 
\begin{equation}
	\begin{array}{l}
		-\frac{\partial^{2} u}{\partial x^{2}}-\frac{\partial^{2} u}{\partial y^{2}}+u(u^{2}-1)=f(x,y),\quad(x,y)\in \Omega.
	\end{array}
\end{equation}
Here, we take $u(x,y)=(\sin(5x)+\cos(5x))\cdot(\sin(5y)+\cos(5y))$, so that the source term $f(x,y)$ can be fabricated by auto-differentiation.

\begin{figure}[htbp]
	\centering
	\includegraphics[width=0.8\textwidth]{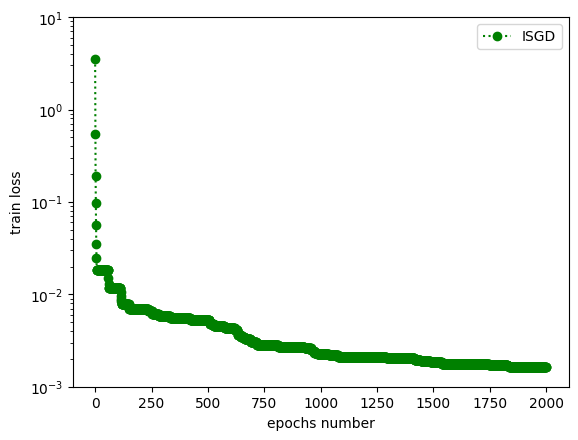}
	\caption{Training loss for the Allen-Cahn equation.}
	\label{loss-ac}
\end{figure}

\begin{figure}[htbp]
	\centering
	\includegraphics[width=\textwidth]{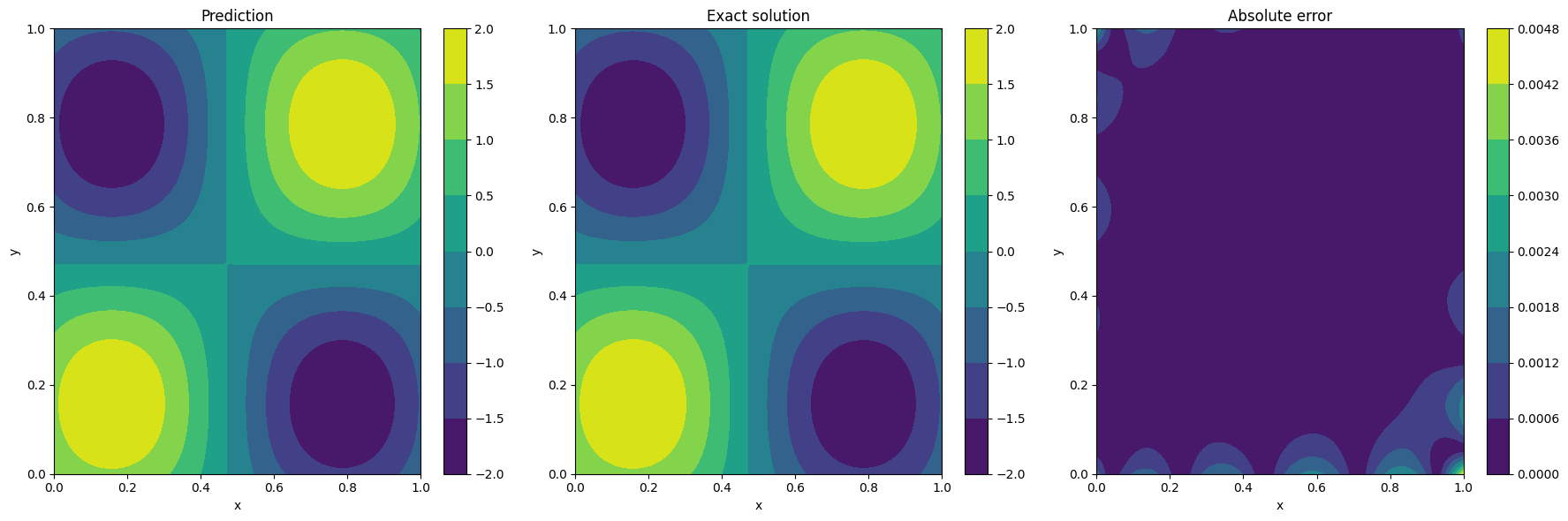}
	\caption{IGD Prediction and Analysis for Allen-Cahn Equation.}
	\label{comparison-ac}
\end{figure}

We sample $N_{b} = 400$ random points along the boundary $\partial \Omega$ and $N_{f} = 2,000$ random points within the domain $\Omega$. For all computations, we employ a neural network architecture consisting of $6$ hidden layers, each with 128 neurons and $tanh$ activation functions. The model is trained over 2,000 epochs using a learning rate of \(lr = 0.5\), resulting in a relative $L^{2}$ error of $0.033 \%$. As depicted in Figure~\ref{loss-ac}, the IGD method demonstrates remarkable accuracy with a small number of training iterations. Figure~\ref{comparison-ac} showcases the solution predicted for the Allen-Cahn equation, alongside the exact solution and the corresponding absolute error. The strong agreement between the predicted and exact results underscores the efficiency and reliability of our proposed method.

To better align the theoretical analysis with the experiments, we train shallow neural networks with a single hidden layer of 128 neurons.
The relative $L^{2}$ error is $0.04\%$, show that the IGD algorithm still performs well under the shallow setting, which is consistent with our theoretical analysis.

\subsection{Computational Efficiency Comparison}
In this section, we show that our IGD algorithm can efficiently solve multi-scale problems compared to the commonly used first order gradient descent optimizers such as SGD and Adam.
For the Poisson equation, the Helmholtz equation and the Allen-Cahn equation, we choose $u(x, y) = \sin(\pi x) \sin(\pi y) +0.1 \sin(10 \pi x) \sin(10 \pi y)$ as the final multi-scale solution, and report the total number of gradient descent iterations \#Iterations and the final relative $L^{2}$ errors for different optimizers.

In fact, the computational efficiency of our IGD method has beed stated in Table 1 of our previous paper \cite{23}, where we have compared the performance of SGD optimizer, Adam optimizer and our IGD optimizer in training PINNs to solve different differential equations.
Like in \cite{23}, we applied a practical ``IGD,Adam'' optimizer for the PINNs training with multi-scale solutions.
Here ``IGD,Adam'' means that we first use IGD with large learning rate for a certain number of iterations, and then switch to Adam with small learning rate to fine tune the convergence result of IGD. In the sub-optimization problem, we also apply a outer Adam optimizer to compute $\theta_{n+1}$.
For our ``IGD,Adam'' algorithm, we note \#Iterations = ($K_0\cdot K_1+K_2)\cdot batchs$, where $K_0$ is the maximum iterations for the IGD, and $K_1$ is the maximum Adam iterations for every sub-optimization problem, and $K_2$ is the maximum iterations for the outer Adam optimizer to fine tuning the convergence of the IGD method.
The wall-clock computational time is proportional to \#Iterations, so the computational efficiency can be compared by the \#Iterations.

As shown in Table \ref{compare-time}, while standard optimizers like SGD and Adam require many epochs to converge on multi-scale problems, our IGD approach achieves accurate solutions efficiently, with a comparable computational burden.

\begin{table}[tbp]
	\centering
	\caption{Hyper-parameters used in the three optimizers for different equations with multi-scale solution $u(x, y) = \sin(\pi x) \sin(\pi y) +0.1 \sin(10 \pi x) \sin(10 \pi y)$.  ``IGD, Adam" is referred as Algorithm 1 in \cite{23}.}\label{compare-time}
	\begin{tabular}{|c|c|c|c|c|}\hline
		Example & Optimizer & Learning rate & \#Iterations & $L^2$ Error \\\hline
		\multirow{3}*{Poisson Equation} & SGD & 0.0005 & 2,000,000 & $0.71\%$ \\\cline{2-5}
		& Adam & 0.0005 & 2,000,000 & $1.93\%$ \\\cline{2-5}
		& \textbf{IGD, Adam} & 0.5, 0.0005 & \textbf{1,100,000} & $\textbf{0.50\%}$ \\\hline
		\multirow{3}*{Helmholtz Equation} & SGD & 0.0005 & 5,000,000 & $0.88\%$ \\\cline{2-5}
		& Adam & 0.0005 & 5,000,000 & $0.78\%$ \\\cline{2-5}
		& \textbf{IGD, Adam} & 0.5, 0.0005 & \textbf{2,200,000} & $\textbf{0.20\%}$ \\\hline
		\multirow{3}*{Allen-Cahn Equation} & SGD & 0.0005 & 5,000,000 & $0.43\%$ \\\cline{2-5}
		& Adam & 0.0005 & 5,000,000 & $0.39\%$ \\\cline{2-5}
		& \textbf{IGD, Adam} & 0.5, 0.0005 & \textbf{2,200,000} & $\textbf{0.18\%}$ \\\hline
	\end{tabular}
\end{table}

\subsection{Learning Rate Comparison}
In this section, we conduct numerical experiments across various learning rate scales for different optimizers.
While SGD and Adam are known to converge only for small learning rate, Theorem \ref{th-lr} indicate that our IGD optimizer can choose the learning rate arbitrarily.

We report the learning rate value from 0.5 to 0.0005. For vary large learning rate, it will be difficult and time-consuming to solve the sub-optimization problem (2.9), so we limit the largest learning rate of IGD to 0.5 in practice.
We train 2,000 epochs for all the learning rates, except 20,000 epochs for the smallest learning rate 0.0005.
The experimental results are listed in the Table~\ref{table_lr}. We observe that the IGD method maintains stable and efficient convergence regardless of whether the learning rate is large or small. These  results provide stronger empirical support for our theoretical claims regarding the robustness and stability of IGD.

\begin{table}[tbp]
	\centering
	\caption{Relative $L^{2}$ errors of different optimizers with different learning rates. ``nan'' means training failed.}\label{table_lr}
	\begin{tabular}{|c|c|c|c|c|c|}\hline
		Example & Learning rate & 0.5 & 0.05 & 0.005 & 0.0005 \\\hline
		\multirow{3}*{Poisson Equation} & SGD $L^2$ Error & nan & nan & $41.71\%$ & $1.37\%$ \\\cline{2-6}
		& Adam $L^2$ Error & $115.07\%$  & $31.08  \%$  & $2.52 \%$  & \textbf{0.36\%} \\\cline{2-6}
		& \textbf{IGD $L^2$ Error} &  \textbf{0.31\%}  & \textbf{0.51\%}  & \textbf{0.55\%}  & $0.75\%$  \\\hline
		\multirow{3}*{Helmholtz Equation} & SGD $L^2$ Error & nan & nan & $74.42\%$ & $6.41\%$ \\\cline{2-6}
		& Adam $L^2$ Error & $101.36\%$  & $166.27\%$  & $11.65\%$  & \textbf{0.88 \%} \\\cline{2-6}
		& \textbf{IGD $L^2$ Error} & \textbf{0.17\%}  & \textbf{0.33\%}  & \textbf{1.90\%}  & $1.83\%$  \\\hline
		\multirow{3}*{Allen-Cahn Equation} & SGD $L^2$ Error & nan & nan & $13.85\%$ & $2.11\%$ \\\cline{2-6}
		& Adam $L^2$ Error & $101.11\%$  & $11.91 \%$  & $2.49 \%$  & \textbf{0.28 \%} \\\cline{2-6}
		& \textbf{IGD $L^2$ Error} & \textbf{0.04\%} & \textbf{0.05\%} & \textbf{0.23\%} & $0.50\%$  \\\hline
	\end{tabular}
\end{table}

\section{Conclusion and Discussion}
In this paper, we have demonstrated the convergence of implicit gradient descent in training two-layer PINNs under suitable over-parameterization and mild assumptions on the training samples. Apart from the class of PDEs we are considering, the methods used in this paper can be naturally extended to a broader class of PDEs. Compared to GD in training PINNs, IGD imposes no requirements on the learning rate due to the implicit updating rule, while GD requires the learning rate $\eta=\mathcal{O}(\lambda_0)$. Due to the dependence on the unknown $\lambda_0$, we must choose a small enough learning rate $\eta$ when using GD for PINNs. However, choosing such a small learning rate can lead to slow convergence. 

Despite the impressive convergence results of IGD, there are some limitations in this work. For instance, we have not taken into account the error associated with the sub-optimization problem during the implicit updating process, which would be interesting to investigate in future work. Moreover, exploring more efficient optimization algorithms for sub-optimization problems is also part of future work, as this can make empirical results closer to theoretical results. As for the generalization bounds for the trained neural networks, it is not difficult to extend the result in \cite{36} to IGD for certain regression or classification problems. However, the generalization of trained PINNs remains an open problem, which is left for future investigation.

\section*{Acknowledgments}
Z.Y. Huang was partially supported by NSFC Projects No. 12025104, 81930119. Y. Li
was partially supported by NSFC Projects No. 62106103 and Fundamental Research Funds for the Central Universities No. ILF240021A24.

\section{Appendix}

\subsection{Proof of Lemma 3.1}
\begin{proof}
From the updating rule of implicit gradient descent (IGD), i.e., $\bm{w}(k+1)=\bm{w}(k)-\eta \nabla L(\bm{w}(k+1))$, we have
\begin{align*}
&s_p(\bm{w}(k+1))-s_p(\bm{w}(k))\\
&=s_p(\bm{w}(k+1))-s_p\left(\bm{w}(k+1)+\eta \nabla L(\bm{w}(k+1))\right)\\
&=-\int_{0}^{\eta} \langle \nabla L(\bm{w}(k+1)), \nabla s_p\left(\bm{w}(k+1)+\alpha\nabla L(\bm{w}(k+1))\right) \rangle d\alpha \\
&= -\int_{0}^{\eta} \langle \nabla L(\bm{w}(k+1)),\nabla s_p(\bm{w}(k+1)) \rangle \\
&\quad -\int_{0}^{\eta} \langle \nabla L(\bm{w}(k+1)), \nabla s_p(\bm{w}(k+1)+\alpha\nabla L(\bm{w}(k+1)))-\nabla s_p(\bm{w}(k+1)) \rangle d\alpha\\
&= -\eta \langle \nabla L(\bm{w}(k+1)), \nabla s_p(\bm{w}(k+1)) \rangle \\
&\quad-\int_{0}^{\eta} \langle \nabla L(\bm{w}(k+1)), \nabla s_p(\bm{w}(k+1)+\alpha\nabla L(\bm{w}(k+1)))-\nabla s_p(\bm{w}(k+1)) \rangle d\alpha\\
&:=\bm{I}_0^p(k+1)-\bm{I}_1^p(k+1),
\end{align*}
where the second equality follows from the Newton-Leibniz formula.

Similarly, for the residual of the boundary, we have
\begin{align*}
&h_j(\bm{w}(k+1))-h_j(\bm{w}(k))\\
&= -\eta \langle \nabla L(\bm{w}(k+1)), \nabla h_j(\bm{w}(k+1)) \rangle \\
&\quad-\int_{0}^{\eta} \langle \nabla L(\bm{w}(k+1)), \nabla h_j(\bm{w}(k+1)+\alpha\nabla L(\bm{w}(k+1)))-\nabla h_j(\bm{w}(k+1)) \rangle d\alpha\\
&:=\bm{I}_0^{n_1+j}(k+1)-\bm{I}_2^j(k+1).
\end{align*}
We denote $\bm{I}_0(k+1)$ as the main term, which is defined as
\[\bm{I}_0(k+1)=(\bm{I}_0^1(k+1),\cdots, \bm{I}_0^{n_1}(k+1),\bm{I}_0^{n_1+1}(k+1),\cdots, \bm{I}_0^{n_1+n_2}(k+1))^T \in \mathbb{R}^{n_1+n_2},\]
and $\bm{I}_1(k+1)$ and $\bm{I}_2(k+1)$ as residual terms, which are defined as 
\[\bm{I}_1(k+1)=(\bm{I}_1^1(k+1),\cdots, \bm{I}_1^{n_1}(k+1))^T\in \mathbb{R}^{n_1} \]
and
\[\bm{I}_2(k+1)=(\bm{I}_2^1(k+1),\cdots, \bm{I}_2^{n_2}(k+1))^T \in \mathbb{R}^{n_2}\]
respectively.

For $\bm{I}_0^p(k+1), p\in [n_1]$, we can derive that
\begin{equation}
\begin{aligned}
&\bm{I}_0^p(k+1)\\
&= -\eta \langle \nabla L(\bm{w}(k+1)), \nabla s_p(\bm{w}(k+1)) \rangle  \\
&= -\eta \sum\limits_{r=1}^m \left\langle \frac{\partial L(\bm{w}(k+1))}{\partial \bm{w}_r}, \frac{\partial s_p(\bm{w}(k+1))}{\partial \bm{w}_r} \right\rangle \\
&=  -\eta \sum\limits_{r=1}^m \left\langle \sum\limits_{i=1}^{n_1}s_i(\bm{w}(k+1))\frac{\partial s_i(\bm{w}(k+1))}{\partial \bm{w}_r}+\sum\limits_{j=1}^{n_2}h_j(\bm{w}(k+1))\frac{\partial h_j(\bm{w}(k+1))}{\partial \bm{w}_r} ,\frac{\partial s_p(\bm{w}(k+1))}{\partial \bm{w}_r} \right\rangle \\
&= -\eta \sum\limits_{i=1}^{n_1}\left(\sum\limits_{r=1}^m \left\langle \frac{\partial s_p(\bm{w}(k+1))}{\partial \bm{w}_r}, \frac{\partial s_i(\bm{w}(k+1))}{\partial \bm{w}_r} \right\rangle\right) s_i(\bm{w}(k+1))\\
&\quad-\eta \sum\limits_{j=1}^{n_2}\left(\sum\limits_{r=1}^m \left\langle \frac{\partial s_p(\bm{w}(k+1))}{\partial \bm{w}_r}, \frac{\partial h_j(\bm{w}(k+1))}{\partial \bm{w}_r} \right\rangle\right) h_j(\bm{w}(k+1))\\
&= -\eta [\bm{G}(k+1)]_p \begin{bmatrix} \bm{s}(\bm{w}(k+1))\\\bm{h}(\bm{w}(k+1)) \end{bmatrix},
\end{aligned}
\end{equation}
where $[\bm{G}(k+1)]_p$ denotes the $p$-th row of $\bm{G}(k+1)$.

Similarly, we can deduce that for $j\in[n_2]$,
\begin{equation}
\bm{I}_0^{n_1+j}(k+1)=-\eta [\bm{G}(k+1)]_{n_1+j} \begin{bmatrix} \bm{s}(\bm{w}(k+1))\\\bm{h}(\bm{w}(k+1)) \end{bmatrix}.
\end{equation}

Combining (6.1) and (6.2) for $p\in [n_1]$ and $j\in [n_2]$ yields
\[\bm{I}_0(k+1)=-\eta \bm{G}(k+1)\begin{bmatrix} \bm{s}(\bm{w}(k+1))\\\bm{h}(\bm{w}(k+1)) \end{bmatrix}. \]
Therefore, we have that
\begin{align*}
\begin{bmatrix} \bm{s}(\bm{w}(k+1))-\bm{s}(\bm{w}(k))\\\bm{h}(\bm{w}(k+1))-\bm{h}(\bm{w}(k)) \end{bmatrix} &= \bm{I}_0(k+1) -\begin{bmatrix} \bm{I}_1(k+1)\\\bm{I}_2(k+1) \end{bmatrix} \\
&=-\eta \bm{G}(k+1) \begin{bmatrix} \bm{s}(\bm{w}(k+1))\\\bm{h}(\bm{w}(k+1)) \end{bmatrix}-\begin{bmatrix} \bm{I}_1(k+1)\\\bm{I}_2(k+1) \end{bmatrix}.
\end{align*}
Finally, a simple vector transformation leads to the conclusion
\begin{equation}
\begin{bmatrix} \bm{s}(\bm{w}(k+1))\\\bm{h}(\bm{w}(k+1)) \end{bmatrix}=(\bm{I}+\eta \bm{G}(k+1))^{-1}  \left( \begin{bmatrix} \bm{s}(\bm{w}(k))\\\bm{h}(\bm{w}(k)) \end{bmatrix} - \begin{bmatrix} \bm{I}_1(k+1)\\\bm{I}_2(k+1) \end{bmatrix}\right).
\end{equation}

\end{proof}

For the sake of readability, we present the concentration inequality of sum of independent sub-Weibull random variables and some preliminaries about the Orlicz norm before the proof of the remaining conclusions.

Recall that for given random variable $X$, the $\psi_{\alpha}$ norm of $X$ is given by 
\[\|X\|_{\psi_{\alpha} }:=\inf\left\{t>0: \mathbb{E}\left[\psi_{\alpha}\left( \frac{|X|}{t} \right) \right] \leq 1 \right\},\]
where $\psi_{\alpha} (x)=e^{x^{\alpha}}-1$ for $x\geq 0$.

Thus, $\|\cdot \|_{\psi_{\alpha}}$ is a norm for $\alpha\geq 1$ and a quasi-norm for $0< \alpha <1$. Moreover, 
\[\|X+Y\|_{\psi_{\alpha} } \leq 2^{1/\alpha}(\|X\|_{\psi_{\alpha} }+\|Y\|_{\psi_{\alpha} }).\]
In the following, we may frequently use the following conclusion that for $X\sim \mathcal{N}(0,1)$, we have 
\[\|X\|_{\psi_2}\leq C, \|X^2\|_{\psi_1} = \|X\|_{\psi_2}^2  \leq C^2, \|X^4\|_{\psi_{\frac{1}{2}}} \leq C^4, \]
where $C$ is a universal constant.

\begin{lemma}[Theorem 3.1 in \cite{24}]
If $X_1,\cdots, X_n$ are independent mean zero random variables with $\|X_i\|_{\psi_{\alpha}}<\infty$ for all $1\leq i\leq n$ and some $\alpha>0$, then for any vector $a=(a_1,\cdots,a_n)\in \mathbb{R}^n$, the following holds true:
\[P\left(\left|\sum\limits_{i=1}^n a_i X_i \right|\geq 2eC(\alpha)\|b\|_2\sqrt{t}+2eL_n^{*}(\alpha)t^{1/\alpha}\|b\|_{\beta(\alpha)}  \right)\leq 2e^{-t}, \ for \ all \ t\geq 0,\]
where $b=(a_1\|X_1\|_{\psi_{\alpha}},\cdots, a_n\|X_n\|_{\psi_{\alpha}}) \in \mathbb{R}^n$,
\begin{equation*}
C(\alpha):=\max\{\sqrt{2},2^{1/\alpha}\}\left\{
\begin{aligned}
\sqrt{8}(2\pi)^{1/4}e^{1/24}(e^{2/e}/\alpha)^{1/\alpha} & , & if \ \alpha<1, \\
4e+2(\log 2)^{1/\alpha} & , & if \ \alpha\geq 1.
\end{aligned}
\right.
\end{equation*}
and for $\beta(\alpha)=\infty$ when $\alpha \leq 1$ and $\beta(\alpha)=\alpha/(\alpha-1)$ when $\alpha>1$,
\begin{equation*}
L_n(\alpha):=\frac{4^{1/\alpha}}{\sqrt{2}\|b\|_2}\times
\left\{
\begin{aligned}
&\|b\|_{\beta(\alpha)} , & if \ \alpha<1, \\
&4e \|b\|_{\beta(\alpha)}/C(\alpha), & if \alpha \geq 1.
\end{aligned}
\right.
\end{equation*}
and $L^{*}_n(\alpha)=L_n(\alpha)C(\alpha)\|b\|_2/\|b\|_{\beta(\alpha)}$.
\end{lemma}

\subsection{Proof of Lemma 3.3}
\begin{proof}
Recall that
\[ \frac{\partial s_p(\bm{w})}{\partial \bm{w}_r}=\frac{a_r}{\sqrt{mn_1}}\left(\sigma^{''}(\bm{w}_r^T\bm{x}_p)w_{r0} \bm{x}_p+\sigma^{'}(\bm{w}_r^T \bm{x}_p) \begin{pmatrix}1\\\bm{0}_d \end{pmatrix}-\sigma^{'''}(\bm{w}_r^T \bm{x}_p)\|\bm{w}_{r1}\|_2^2 \bm{x}_p-2\sigma^{''}(\bm{w}_r^T \bm{x}_p) \begin{pmatrix}0\\\bm{w}_{r1} \end{pmatrix}\right)\]
and
\[\frac{\partial h_j(\bm{w})}{\partial \bm{w}_r}=\frac{a_r }{\sqrt{mn_2}}\sigma^{'}(\bm{w}_r^T \bm{y}_j)\bm{y}_j.\]
Since $\|\bm{G}(0)-\bm{G}^{\infty}\|_2\leq \|\bm{G}(0)-\bm{G}^{\infty}\|_F$, it suffices to bound each entry of $\bm{G}(0)-\bm{G}^{\infty}$, which is of the form
\begin{equation}
\sum\limits_{r=1}^m \left\langle \frac{\partial s_p(\bm{w})}{\partial \bm{w}_r}, \frac{\partial s_j(\bm{w})}{\partial \bm{w}_r}\right\rangle -\mathbb{E}_{\bm{w}} \sum\limits_{r=1}^m \left\langle \frac{\partial s_p(\bm{w})}{\partial \bm{w}_r}, \frac{\partial s_j(\bm{w})}{\partial \bm{w}_r}\right\rangle
\end{equation}
or
\begin{equation}
\sum\limits_{r=1}^m \left\langle \frac{\partial s_p(\bm{w})}{\partial \bm{w}_r}, \frac{\partial h_j(\bm{w})}{\partial \bm{w}_r}\right\rangle -\mathbb{E}_{\bm{w}} \sum\limits_{r=1}^m \left\langle \frac{\partial s_p(\bm{w})}{\partial \bm{w}_r}, \frac{\partial h_j(\bm{w})}{\partial \bm{w}_r}\right\rangle	
\end{equation}
or
\begin{equation}
\sum\limits_{r=1}^m \left\langle \frac{\partial h_p(\bm{w})}{\partial \bm{w}_r}, \frac{\partial h_j(\bm{w})}{\partial \bm{w}_r}\right\rangle -\mathbb{E}_{\bm{w}} \sum\limits_{r=1}^m \left\langle \frac{\partial h_p(\bm{w})}{\partial \bm{w}_r}, \frac{\partial h_j(\bm{w})}{\partial \bm{w}_r}\right\rangle.	
\end{equation}
For the first form (6.4), let
\[Y_r(i)=\sigma^{''}(\bm{w}_r(0)^T\bm{x}_i)w_{r0}(0) \bm{x}_i+\sigma^{'}(\bm{w}_r(0)^T \bm{x}_i) \begin{pmatrix}1\\\bm{0}_d \end{pmatrix}-\sigma^{'''}(\bm{w}_r(0)^T \bm{x}_i)\|\bm{w}_{r1}(0)\|_2^2 \bm{x}_i-2\sigma^{''}(\bm{w}_{r}(0)^T \bm{x}_i) \begin{pmatrix}0\\\bm{w}_{r1}(0) \end{pmatrix}\]
and
\[X_r(ij)=\langle Y_r(i), Y_r(j) \rangle ,\]
then
\[\sum\limits_{r=1}^m \left\langle \frac{\partial s_p(\bm{w})}{\partial \bm{w}_r}, \frac{\partial s_j(\bm{w})}{\partial \bm{w}_r}\right\rangle -\mathbb{E}_{\bm{w}} \sum\limits_{r=1}^m \left\langle \frac{\partial s_p(\bm{w})}{\partial \bm{w}_r}, \frac{\partial s_j(\bm{w})}{\partial \bm{w}_r}\right\rangle=\frac{1}{n_1 m} \sum\limits_{r=1}^m (X_r(ij)-\mathbb{E}X_r(ij)).\]
Note that $|X_r(ij)|\lesssim 1+\|\bm{w}_r(0)\|_2^4$ and  for any random variable $X$, $\|X^2\|_{\psi_{\frac{1}{2}}}=\|X\|_{\psi_1}^2$, thus
\[ \|X_r(ij)\|_{\psi_{\frac{1}{2}}} \lesssim 1+ \left\|\|\bm{w}_r(0)\|_2^4 \right\|_{\psi_{\frac{1}{2}}} \lesssim 1+ \left\|\|\bm{w}_r(0)\|_2^2\right\|_{\psi_1}^2 \lesssim d^2. \]
For the centered random variable, the property of $\psi_{\frac{1}{2}}$ quasi-norm implies that
\begin{equation*}
\|X_r(ij)-\mathbb{E}[X_r(ij)]\|_{\psi_{\frac{1}{2}}} \lesssim \|X_r(ij)\|_{\psi_{\frac{1}{2}}}+\|\mathbb{E}[X_r(ij)]\|_{\psi_{\frac{1}{2}}}\lesssim d^2.
\end{equation*}

Therefore, applying Lemma 6.1 yields that with probability at least $1-\delta$,
\[\left|\sum\limits_{r=1}^m \frac{1}{m}(X_r(ij)-\mathbb{E}[X_r(ij)]) \right| \lesssim \frac{d^2}{\sqrt{m}}\sqrt{\log\left(\frac{1}{\delta}\right)} +\frac{d^2}{m}\left(\log\left(\frac{1}{\delta}\right)\right)^2,\]
which leads directly to that
\begin{equation}
\left|\sum\limits_{r=1}^m \left\langle \frac{\partial s_i(\bm{w})}{\partial \bm{w}_r}, \frac{\partial s_j(\bm{w})}{\partial \bm{w}_r}\right\rangle -\mathbb{E}_{\bm{w}} \sum\limits_{r=1}^m \left\langle \frac{\partial s_i(\bm{w})}{\partial \bm{w}_r}, \frac{\partial s_j(\bm{w})}{\partial \bm{w}_r}\right\rangle\right|\lesssim \frac{d^2}{n_1\sqrt{m}}\sqrt{\log\left(\frac{1}{\delta}\right)} +\frac{d^2}{n_1m}\left(\log\left(\frac{1}{\delta}\right)\right)^2.
\end{equation}

For the second form (6.5), we only need to apply Lemma 6.1 for random variables with finite $\psi_{1}$ norm, which yields that with probability at least $1-\delta$,
\begin{equation}
\left|\sum\limits_{r=1}^m \left\langle \frac{\partial s_i(\bm{w})}{\partial \bm{w}_r}, \frac{\partial h_j(\bm{w})}{\partial \bm{w}_r}\right\rangle -\mathbb{E}_{\bm{w}} \sum\limits_{r=1}^m \left\langle \frac{\partial s_i(\bm{w})}{\partial \bm{w}_r}, \frac{\partial h_j(\bm{w})}{\partial \bm{w}_r}\right\rangle\right|\lesssim \frac{d}{\sqrt{n_1n_2}\sqrt{m}}\sqrt{\log\left(\frac{1}{\delta}\right)} +\frac{d}{\sqrt{n_1n_2}m}\log\left(\frac{1}{\delta}\right).	
\end{equation}
Similarly, for the third form (6.6), applying Lemma 6.1 for random variables with finite $\psi_2$ norm suffices. Thus, we have that with probability at least $1-\delta$,
\begin{equation}
\left|\sum\limits_{r=1}^m \left\langle \frac{\partial h_i(\bm{w})}{\partial \bm{w}_r}, \frac{\partial h_j(\bm{w})}{\partial \bm{w}_r}\right\rangle -\mathbb{E}_{\bm{w}} \sum\limits_{r=1}^m \left\langle \frac{\partial h_i(\bm{w})}{\partial \bm{w}_r}, \frac{\partial h_j(\bm{w})}{\partial \bm{w}_r}\right\rangle\right|\lesssim \frac{1}{n_2\sqrt{m}}\sqrt{\log\left(\frac{1}{\delta}\right)} .
\end{equation}
Combining above results (6.7), (6.8) and (6.9) for the three forms (6.4), (6.5) and (6.6), we can deduce that with probability at least $1-\delta$,
\begin{align*}
&\|\bm{G}(0)-\bm{G}^{\infty}\|_2^2 \\
&\leq \|\bm{G}(0)-\bm{G}^{\infty}\|_F^2 \\
&\lesssim \frac{d^4}{m}\log\left(\frac{n_1+n_2}{\delta}\right)+\frac{d^4}{m^2}\left(\log\left(\frac{n_1+n_2}{\delta}\right)\right)^4 \\
&\lesssim \frac{d^4}{m}\log\left(\frac{n_1+n_2}{\delta}\right),
\end{align*}
where we omit the high-order term about $1/m$ and $\log((n_1+n_2)/\delta)$.

Thus when $ \sqrt{\frac{d^4}{m}\log\left(\frac{n_1+n_2}{\delta}\right)}\lesssim \frac{\lambda_0}{4}$, i.e.,
\[ m=\Omega\left(\frac{d^4}{\lambda_0^2}\log\left(\frac{n_1+n_2}{\delta}\right)\right),\]
we have $\lambda_{min}(\bm{G}(0))\geq \frac{3}{4}\lambda_0 $.

\end{proof}

\subsection{Proof of Lemma 3.4}
\begin{proof}
As $\|\bm{G}(\bm{w})-\bm{G}(0)\|_2\leq \|\bm{G}(\bm{w})-\bm{G}(0)\|_F$, it suffices to bound each entry of $|\bm{G}(\bm{w})-\bm{G}(0)|$.

For $i\in[1,n_1]$ and $j\in[1,n_1]$, we have
\begin{align*}
&|G_{ij}(\bm{w})-G_{ij}(0)|\\
&= \left|\sum\limits_{r=1}^m \left\langle \frac{\partial s_i(\bm{w})}{\partial \bm{w}_r}, \frac{\partial s_j(\bm{w})}{\partial \bm{w}_r}\right\rangle-\left\langle \frac{\partial s_i(\bm{w}(0))}{\partial \bm{w}_r}, \frac{\partial s_j(\bm{w}(0))}{\partial \bm{w}_r}\right\rangle \right|\\
&\lesssim R\frac{1}{n_1}\frac{1}{m}\sum\limits_{r=1}^m(\|\bm{w}_r(0)\|_2^4+\|\bm{w}_r(0)\|_2^3+\|\bm{w}_r(0)\|_2^2+\|\bm{w}_r(0)\|_2+1) \\
&\lesssim R\frac{1}{n_1}\frac{1}{m}\sum\limits_{r=1}^m(\|\bm{w}_r(0)\|_2^4+1),
\end{align*}
where the first inequality is due to $R\leq 1$ and the second inequality follows from the Young's inequality.

For $i\in[1,n_1]$ and $j\in[n_1+1,n_1+n_2]$, we have
\begin{align*}
&|G_{ij}(\bm{w})-G_{ij}(0)|\\
&= \left|\sum\limits_{r=1}^m \left\langle \frac{\partial s_i(\bm{w})}{\partial \bm{w}_r}, \frac{\partial h_j(\bm{w})}{\partial \bm{w}_r}\right\rangle-\left\langle \frac{\partial s_i(\bm{w}(0))}{\partial \bm{w}_r}, \frac{\partial h_j(\bm{w}(0))}{\partial \bm{w}_r}\right\rangle \right|\\
&\lesssim R\frac{1}{\sqrt{n_1n_2}}\frac{1}{m}\sum\limits_{r=1}^m(\|\bm{w}_r(0)\|_2^2+\|\bm{w}_r(0)\|_2+1) \\
&\lesssim R\frac{1}{\sqrt{n_1n_2}}\frac{1}{m}\sum\limits_{r=1}^m(\|\bm{w}_r(0)\|_2^2+1).
\end{align*}

For $i\in[n_1+1,n_1+n_2]$ and $j\in[n_1+1,n_1+n_2]$, we have
\begin{align*}
&|G_{ij}(\bm{w})-G_{ij}(0)|\\
&= \left|\sum\limits_{r=1}^m \left\langle \frac{\partial h_i(\bm{w})}{\partial \bm{w}_r}, \frac{\partial h_j(\bm{w})}{\partial \bm{w}_r}\right\rangle-\left\langle \frac{\partial h_i(\bm{w}(0))}{\partial \bm{w}_r}, \frac{\partial h_j(\bm{w}(0))}{\partial \bm{w}_r}\right\rangle \right|\\
&\lesssim R\frac{1}{n_2}\frac{1}{m}.
\end{align*}

Combining above results yields that
\begin{align*}
&\|\bm{G}(\bm{w})-\bm{G}(0)\|_2^2 \\
&\leq \|\bm{G}(\bm{w})-\bm{G}(0)\|_F^2\\
&\lesssim R^2+R^2 \left(\frac{1 }{m}\sum\limits_{r=1}^m \|\bm{w}_r(0)\|_2^4\right)^2+R^2  \left(\frac{1 }{m}\sum\limits_{r=1}^m \|\bm{w}_r(0)\|_2^2\right)^2.
\end{align*}
For the second term, applying Lemma 6.1 for random variable $\|\bm{w}_r(0)\|_2^4-\mathbb{E}[\|\bm{w}_r(0)\|_2^4]$ with $a_r=1/m$ and $\alpha=1/2$ yields that with probability at least $1-\delta$,
\[\frac{1 }{m}\sum\limits_{r=1}^m \|\bm{w}_r(0)\|_2^4 \lesssim d^2+\frac{d^2}{\sqrt{m}}\sqrt{\log\left(\frac{1}{\delta}\right)}+\frac{d^2}{m}\left(\log\left(\frac{1}{\delta}\right)\right)^2.\]
Similarly, for the third term, we have that with probability at least $1-\delta$,
\[\frac{1 }{m}\sum\limits_{r=1}^m \|\bm{w}_r(0)\|_2^2 \lesssim d+\frac{d}{\sqrt{m}}\sqrt{\log\left(\frac{1}{\delta}\right)}+\frac{d}{m}\log\left(\frac{1}{\delta}\right).\]
Finally, we can deduce that with probability at least $1-\delta$,
\[\|\bm{G}(\bm{w})-\bm{G}(0)\|_2 \leq CR\max\left(d^2,\frac{d^2}{\sqrt{m}}\sqrt{\log\left(\frac{1}{\delta}\right)}, \frac{d^2}{m}\left(\log\left(\frac{1}{\delta}\right)\right)^2 \right),\]
where $C$ is a universal constant.

\end{proof}

\subsection{Proof of Lemma 3.5}
\begin{proof}
Applying Lemma 6.1 for random variable $X_i=w_{ri}(0)^2-\mathbb{E}[w_{ri}(0)^2]$ with $a_i=1$ and $\alpha=1$, we can deduce that for fixed $r\in [m]$,
\[\|\bm{w}_r(0)\|_2^2 \leq C\left(d+\sqrt{d\log\left(\frac{1}{\delta}\right)}+\log\left(\frac{1}{\delta}\right) \right)\].

Therefore, the following holds with probability at least $1-\delta$.
\[\|\bm{w}_r(0)\|_2^2 \leq C\left(d+\sqrt{d\log\left(\frac{m}{\delta}\right)}+\log\left(\frac{m}{\delta}\right) \right), \quad \text{for} \ \forall r\in[m].\]	
\end{proof}

\subsection{Proof of Lemma 3.6}
\begin{proof}
For the initial value of PINNs, we have 
\begin{equation}
\begin{aligned}
L(0)&=\frac{1}{2}\sum\limits_{p=1}^{n_1}s_p^2(\bm{w}(0))+\frac{1}{2}\sum\limits_{j=1}^{n_2}h_j^2(\bm{w}(0)) \\
&= \frac{1}{2n_1}\sum\limits_{p=1}^{n_1} \left(\frac{1}{\sqrt{m}} \sum\limits_{r=1}^m a_r(0)[\sigma^{'}(\bm{w}_r(0)^T \bm{x}_p)w_{r0}(0) -\sigma^{''}(\bm{w}_r(0)^T\bm{x}_p )\|\bm{w}_{r1}(0)\|_2^2]-f(\bm{x}_p) \right)^2\\
&\quad + \frac{1}{2n_2}\sum\limits_{j=1}^{n_2} \left(\frac{1}{\sqrt{m}} \sum\limits_{r=1}^m a_r(0)\sigma(\bm{w}_r(0)^T \bm{y}_j)-g(\bm{y}_j)\right)^2\\
&\leq \frac{1}{n_1}\sum\limits_{p=1}^{n_1} \left(\frac{1}{\sqrt{m}} \sum\limits_{r=1}^m a_r(0)[\sigma^{'}(\bm{w}_r(0)^T \bm{x}_p)w_{r0}(0) -\sigma^{''}(\bm{w}_r(0)^Tx_p )\|\bm{w}_{r1}(0)\|_2^2]\right)^2+(f(\bm{x}_p) )^2\\
&\quad + \frac{1}{n_2}\sum\limits_{j=1}^{n_2} \left(\frac{1}{\sqrt{m}} \sum\limits_{r=1}^m a_r(0)\sigma(\bm{w}_r(0)^T \bm{y}_j)\right)^2+(g(\bm{y}_j))^2.
\end{aligned}
\end{equation}
For the first term in (6.10), note that $\mathbb{E}\left[a_r(0)\left[\sigma^{'}(\bm{w}_r(0)^T \bm{x}_p)w_{r0}(0) -\sigma^{''}(\bm{w}_r(0)^T\bm{x}_p )\|\bm{w}_{r1}(0)\|_2^2\right]\right]=0$ and
\begin{align*}
&\left|a_r(0)\left[\sigma^{'}(\bm{w}_r(0)^T \bm{x}_p)w_{r0}(0) -\sigma^{''}(\bm{w}_r(0)^T\bm{x}_p )\|\bm{w}_{r1}(0)\|_2^2\right]\right|\\
&\lesssim |w_{r0}(0)|+\|\bm{w}_{r1}(0)\|_2^2 \\
&\lesssim  1+ |w_{r0}(0)|^2+\|\bm{w}_{r1}(0)\|_2^2 \\
&= 1+\|\bm{w}_{r}(0)\|_2^2.
\end{align*}
Therefore,
\[\left\|  a_r(0)\left[\sigma^{'}(\bm{w}_r(0)^T \bm{x}_p)w_{r0}(0) -\sigma^{''}(\bm{w}_r(0)^T\bm{x}_p )\|\bm{w}_{r1}(0)\|_2^2\right] \right\|_{\psi_1} \leq Cd.\]
Let $X_r=a_r(0)\left[\sigma^{'}(\bm{w}_r(0)^T \bm{x}_p)w_{r0}(0) -\sigma^{''}(\bm{w}_r(0)^T\bm{x}_p )\|\bm{w}_{r1}(0)\|_2^2\right]$, then Lemma 6.1 implies that with probability at least $1-\delta$,
\[\left|\sum\limits_{r=1}^m \frac{X_r}{\sqrt{m}}\right|\lesssim d\sqrt{\log\left(\frac{1}{\delta}\right)}+\frac{d}{\sqrt{m}} \log\left(\frac{1}{\delta}\right). \]
For the second term in (6.10), we have
\[\frac{1}{\sqrt{m}} \sum\limits_{r=1}^m a_r(0)\sigma(w_r(0)^T \bm{y}_j)=\frac{1}{\sqrt{m}} \sum\limits_{r=1}^m a_r(0)(\sigma(\bm{w}_r(0)^T \bm{y}_j)-\sigma(0))+a_r(0)\sigma(0),\]
note that $\mathbb{E}[a_r(0)(\sigma(\bm{w}_r(0)^T \bm{y}_j)-\sigma(0))]=0$ and
\[|a_r(0)(\sigma(\bm{w}_r(0)^T \bm{y}_j)-\sigma(0))|\lesssim |\bm{w}_r(0)^T \bm{y}_j|,\]
thus
\[ \| a_r(0)(\sigma(\bm{w}_r(0)^T \bm{y}_j)-\sigma(0))\|_{\psi_2} \leq C,\]
as $\bm{w}_r(0)^T \bm{y}_j\sim N(0,\|\bm{y}_j\|_2^2)$.

Let $Y_r=a_r(0)(\sigma(\bm{w}_r(0)^T \bm{y}_j)-\sigma(0))$, applying Lemma 6.1 yields that with probability at least $1-\delta$,
\[\left|\sum\limits_{r=1}^m \frac{Y_r}{\sqrt{m}}\right|\lesssim \sqrt{\log\left(\frac{1}{\delta}\right)} + \frac{1}{\sqrt{m}} \log\left(\frac{1}{\delta}\right).\]

It remains only to handle $\frac{1}{\sqrt{m}}\sum_{r=1}^m a_r(0)\sigma(0)$, which is sub-Gaussian. Specifically, for any
$\lambda \in \mathbb{R}$, we have
\begin{align*}
&\mathbb{E} \exp\left(\frac{\lambda}{\sqrt{m}}\sum_{r=1}^m a_r(0)\sigma(0)\right) \\
&= \prod_{r=1}^{m} \mathbb{E} \exp\left(\frac{\lambda}{\sqrt{m}} a_r(0)\sigma(0)\right) \\
&\leq \prod_{r=1}^{m} \mathbb{E} \exp\left( \frac{\lambda^2}{2m}\sigma(0)^2 \right)\\
&= \mathbb{E} \exp\left( \frac{\lambda^2}{2}\sigma(0)^2 \right),
\end{align*}
where the first inequality follows from the fact that the Rademacher variable $a_r(0)$ is sub-Gaussian.

Thus with probability at least $1-\delta$,
\[\left|\frac{1}{\sqrt{m}}\sum\limits_{r=1}^m a_r(0)\sigma(0)\right|\lesssim \sqrt{\log\left(\frac{1}{\delta}\right)} .\]
Combining all results above yields that
\[L(0)\lesssim d^2\log\left(\frac{n_1+n_2}{\delta}\right)+\frac{d^2}{m} \left(\log\left(\frac{n_1+n_2}{\delta}\right)\right)^2 \]
holds with probability at least $1-\delta$.

\end{proof}

\subsection{Proof of Lemma 3.7}
\begin{proof}
Recall that
\begin{equation}
\frac{\partial s_p(\bm{w})}{\partial \bm{w}_r}=\frac{a_r}{\sqrt{mn_1}}\left(\sigma^{''}(\bm{w}_r^T\bm{x}_p)w_{r0} \bm{x}_p+\sigma^{'}(\bm{w}_r^T \bm{x}_p) \begin{pmatrix}1\\\bm{0}_d \end{pmatrix}-\sigma^{'''}(\bm{w}_r^T \bm{x}_p)\|\bm{w}_{r1}\|_2^2 \bm{x}_p-2\sigma^{''}(\bm{w}_r^T \bm{x}_p) \begin{pmatrix}0\\\bm{w}_{r1} \end{pmatrix}\right)	
\end{equation}
and
\begin{equation}
\frac{\partial h_j(\bm{w})}{\partial \bm{w}_r}=\frac{a_r}{\sqrt{mn_2}}\sigma^{'}(\bm{w}_r^T \bm{y}_j)\bm{y}_j.	
\end{equation}
Thus, under the setting that Condition 1 holds for $t=0,\cdots, k$, from (6.11) and (6.12), we can deduce that
\[\left\|\frac{\partial s_p(\bm{w}(t))}{\partial \bm{w}_r}\right\|_2\lesssim \frac{B^2}{\sqrt{mn_1}}, \quad \left\|\frac{\partial h_j(\bm{w}(t))}{\partial \bm{w}_r}\right\|_2\lesssim \frac{1}{\sqrt{mn_2}}\]
holds for $t=0,\cdots, k$.

From the updating rule, i.e.,
\[\bm{w}_r(t+1)= \bm{w}_r(t)-\eta \sum\limits_{p=1}^{n_1}s_p(\bm{w}(t+1))\frac{\partial s_p(\bm{w}(t+1))}{\partial \bm{w}_r}-\eta \sum\limits_{j=1}^{n_2}h_j(\bm{w}(t+1))\frac{\partial h_j(\bm{w}(t+1))}{\partial \bm{w}_r},\]
we have that for $t=0,\cdots,k-1$,
\begin{align*}
\|\bm{w}_r(t+1)-\bm{w}_r(t)\|_2&\leq \eta\left\|  \sum\limits_{p=1}^{n_1}s_p(\bm{w}(t+1))\frac{\partial s_p(\bm{w}(t+1))}{\partial \bm{w}_r}+\sum\limits_{j=1}^{n_2}h_j(t+1)\frac{\partial h_j(\bm{w}(t+1))}{\partial \bm{w}_r}    \right \|_2 \\
&\leq \eta\sum\limits_{p=1}^{n_1}|s_p(\bm{w}(t+1))|\left\|\frac{\partial s_p(\bm{w}(t+1))}{\partial \bm{w}_r}\right\|_2+\eta \sum\limits_{j=1}^{n_2}|h_j(\bm{w}(t+1))|\left\|\frac{\partial h_j(\bm{w}(t+1))}{\partial \bm{w}_r}\right\|_2  \\
&\lesssim \eta \sum\limits_{p=1}^{n_1}|s_p(\bm{w}(t+1))|\frac{B^2}{\sqrt{mn_1}}+\eta \sum\limits_{j=1}^{n_2}|h_j(\bm{w}(t+1))|\frac{1}{\sqrt{mn_2}} \\
&\leq \eta  \frac{B^2}{\sqrt{m}}\|\bm{s}(\bm{w}(t+1))\|_2+\eta \frac{1}{\sqrt{m}} \|\bm{h}(\bm{w}(t+1))\|_2 \\
&\lesssim \frac{\eta B^2}{\sqrt{m}} \left\| \begin{bmatrix} \bm{s}(\bm{w}(t+1))\\\bm{h}(\bm{w}(t+1)) \end{bmatrix} \right\|_2\\
&\leq  \frac{\eta B^2}{\sqrt{m}} \left(1+\frac{\eta\lambda_0}{2}\right)^{-\frac{t+1}{2}}\left\| \begin{bmatrix} \bm{s}(0)\\\bm{h}(0) \end{bmatrix} \right\|_2,
\end{align*}
where the second inequality follows from triangle inequality, the third inequality follows from Cauchy's inequality and the last inequality is due to the induction.

Summing $t$ from $0$ to $k-1$ yields that
\begin{equation}
\begin{aligned}
\|\bm{w}_r(k)-\bm{w}_r(0)\|_2&\leq \sum\limits_{t=0}^{k-1} \|\bm{w}_r(t+1)-\bm{w}_r(t)\|_2 \\
&\leq C\frac{\eta B^2}{\sqrt{m}}\sum\limits_{t=0}^{k-1} \left(1+\frac{\eta\lambda_0}{2}\right)^{-\frac{t+1}{2}}\left\| \begin{bmatrix} \bm{s}(0)\\ \bm{h}(0) \end{bmatrix} \right\|_2 \\
&\leq C\frac{\eta B^2}{\sqrt{m}} \sum\limits_{t=1}^{\infty}\left(1+\frac{\eta\lambda_0}{2}\right)^{-\frac{t}{2}}\left\| \begin{bmatrix} \bm{s}(0)\\ \bm{h}(0) \end{bmatrix} \right\|_2\\
&\leq C\frac{\eta B^2}{\sqrt{m}} \frac{1}{\eta \lambda_0}\left\| \begin{bmatrix} \bm{s}(0)\\ \bm{h}(0) \end{bmatrix} \right\|_2\\
&\leq \frac{C B^2}{\sqrt{m}\lambda_0}\left\| \begin{bmatrix} \bm{s}(0)\\ \bm{h}(0) \end{bmatrix} \right\|_2,
\end{aligned}
\end{equation}
where $C$ is a universal constant.

For $t=k$, we can only deduce that
\begin{equation}
\begin{aligned}
&\|\bm{w}_r(k+1)-\bm{w}_r(k)\|_2\\
&\leq \eta  \left\|\sum\limits_{p=1}^{n_1}s_p(\bm{w}(k+1))\frac{\partial s_p(\bm{w}(k+1))}{\partial \bm{w}_r}+ \sum\limits_{j=1}^{n_2}h_j(\bm{w}(k+1))\frac{\partial h_j(\bm{w}(k+1))}{\partial \bm{w}_r}\right\|_2\\
&\leq \eta\frac{1}{\sqrt{mn_1}} (\|\bm{w}_r(k+1)\|_2^2+1)\sum\limits_{p=1}^{n_1} |s_p(\bm{w}(k+1))|+\eta\frac{1}{\sqrt{mn_2}}\sum\limits_{j=1}^{n_2}|h_j(\bm{w}(k+1))|\\
&\lesssim \eta \frac{1}{\sqrt{m}} (\|\bm{w}_r(k+1)\|_2^2+1) (\|\bm{s}(k+1)\|_2+\|\bm{h}(k+1)\|_2) \\
&\lesssim \eta \frac{1}{\sqrt{m}} (\|\bm{w}_r(k+1)\|_2^2+1) \left\|\begin{bmatrix} \bm{s}(k+1)\\ \bm{h}(k+1)\end{bmatrix} \right\|_2 \\
&= \eta \frac{1}{\sqrt{m}} (\|\bm{w}_r(k+1)\|_2^2+1) \sqrt{L(k+1)},
\end{aligned}
\end{equation}
where the third inequality follows from Cauchy's inequality.

Fortunately, the implicit updating rule, i.e.,
\[\bm{w}(k+1)=\mathop{\text{argmin}}_{\bm{w}} \frac{1}{2}\|\bm{w}-\bm{w}(k)\|_2^2+\eta L(\bm{w}),\]
implies that
\[\|\bm{w}(k+1)-\bm{w}(k)\|_2^2 \leq 2\eta (L(k)-L(k+1)),\]
thus we have 
\begin{equation}
\|\bm{w}_r(k+1)-\bm{w}_r(k)\|_2^2 \leq 2\eta L(k), \quad \text{for} \ \forall r\in [m]
\end{equation}
and $L(k+1)\leq L(k)\leq L(0)$. 

Plugging (6.15) into (6.14) yields that
\begin{equation}
\begin{aligned}
&\|\bm{w}_r(k+1)-\bm{w}_r(k)\|_2\\
&\lesssim \eta \frac{1}{\sqrt{m}} (\|\bm{w}_r(k+1)\|_2^2+1) \sqrt{L(k+1)}\\
&\leq \eta \frac{1}{\sqrt{m}} (2\|\bm{w}_r(k+1)-\bm{w}_r(k)\|_2^2+2\|\bm{w}_r(k)\|_2^2+1) \sqrt{L(k+1)} \\
&\lesssim \frac{\eta}{\sqrt{m}} (\eta L(k)+B^2+1)\sqrt{L(0)}\\
&\lesssim \frac{\eta}{\sqrt{m}} (\eta L(0)+B^2)\sqrt{L(0)}.
\end{aligned}
\end{equation}
Combining (6.13) and (6.16), we have
\begin{equation}
\begin{aligned}
\|\bm{w}_r(k+1)-\bm{w}_r(0)\|_2 &\leq \|\bm{w}_r(k+1)-\bm{w}_r(k)\|_2+\|\bm{w}_r(k)-\bm{w}_r(0)\|_2\\
&\leq \frac{C\eta}{\sqrt{m}} (\eta L(0)+B^2)\sqrt{L(0)}+\frac{C B^2}{\sqrt{m}\lambda_0} \sqrt{L(0)},
\end{aligned}
\end{equation}
where $C$ is a universal constant.
\end{proof}

\subsection{Proof of Lemma 3.8}
\begin{proof}
Thanks to Lemma 3.7 that provides an upper bound for $\|\bm{w}_r(k+1)-\bm{w}_r(0)\|_2$, i.e., (6.17), then applying Lemma 3.4 yields that
\begin{align*}
&\|\bm{G}(k+1)-\bm{G}(0)\|_2\\
&\leq \left(C\max\left(d^2,\frac{d^2}{\sqrt{m}}\sqrt{\log\left(\frac{1}{\delta}\right)}, \frac{d^2}{m}\left(\log\left(\frac{1}{\delta}\right)\right)^2 \right)\right) \left(\frac{C\eta}{\sqrt{m}} \left(\eta L(0)+B^2\right)\sqrt{L(0)}+\frac{C B^2}{\sqrt{m}\lambda_0} \sqrt{L(0)}\right) \\
&\leq Cd^2 \frac{\sqrt{L(0)}}{\sqrt{m}} \max\left\{\eta^2L(0), \frac{B^2}{\lambda_0} \right\}\\
&\leq \frac{\lambda_0}{4},
\end{align*}
where the last inequality requires that
\[m\geq C\frac{d^4 L(0)}{\lambda_0^2}\max\left\{\eta^4L^2(0), \frac{B^4}{\lambda_0^2} \right\},\]
i.e.,
\begin{equation}
m\geq C\frac{d^{8}}{\lambda_0^2}\log\left(\frac{n_1+n_2}{\delta}\right) \max\left\{\eta^4 d^2 \log\left(\frac{n_1+n_2}{\delta}\right), \frac{1}{\lambda_0^2} \left(\log\left(\frac{m}{\delta}\right)\right)^2   \right\}  .
\end{equation}
Thus, $\lambda_{min}(\bm{G}(k+1))\geq \frac{\lambda_0}{2}$.

\end{proof}

\subsection{Proof of Lemma 3.9}
\begin{proof}
Recall that
\[\bm{I}_1^p(k+1)=\int_{0}^{\eta} \langle \nabla L(\bm{w}(k+1)), \nabla s_p(\bm{w}(k+1)+\alpha\nabla(\bm{w}(k+1)))-\nabla s_p(\bm{w}(k+1)) \rangle d\alpha\]
and
\[\bm{I}_2^j(k+1)=\int_{0}^{\eta} \langle \nabla L(\bm{w}(k+1)), \nabla h_j(\bm{w}(k+1)+\alpha \nabla L(\bm{w}(k+1)))-\nabla h_j(\bm{w}(k+1)) \rangle d\alpha.\]
Thus for $\bm{I}_1^p(k+1)$, from the updating rule of IGD, we have
\begin{align*}
\bm{I}_1^p(k+1)&= \sum\limits_{r=1}^m \int_{0}^{\eta} \left\langle \frac{\partial L(\bm{w}(k+1))}{\partial \bm{w}_r}, \frac{\partial s_p(\bm{w}(k+1)+\alpha\frac{\partial L(\bm{w}(k+1))}{\partial \bm{w}} )}{\partial \bm{w}_r} -\frac{\partial s_p(\bm{w}(k+1))}{\partial \bm{w}_r} \right\rangle d\alpha.
\end{align*}
Note that
\[\frac{\partial s_p(\bm{w})}{\partial \bm{w}_r}=\frac{1}{\sqrt{n_1}} \frac{a_r}{\sqrt{m}} \left[ \sigma^{''}(\bm{w}_r^T\bm{x}_p)w_{r0} \bm{x}_p+\sigma^{'}(\bm{w}_r^T \bm{x}_p) \begin{pmatrix}1\\0_d \end{pmatrix}-\sigma^{'''}(\bm{w}_r^T \bm{x}_p)\|\bm{w}_{r1}\|_2^2 \bm{x}_p-2\sigma^{''}(\bm{w}_r^T \bm{x}_p) \begin{pmatrix}0\\\bm{w}_{r1} \end{pmatrix} \right],\]
thus for any $\bm{w}=(\bm{w}_1,\cdots,\bm{w}_m)$ and $\hat{\bm{w}}=(\hat{\bm{w}}_1,\cdots, \hat{\bm{w}}_m)$ with $\|\bm{w}_r\|_2\leq B, \|\hat{\bm{w}}_r\|_2\leq B$ for all $r\in[m]$ (as for any $\alpha \in [0,\eta]$, $\bm{w}(k+1)+\alpha\frac{\partial L(\bm{w}(k+1))}{\partial \bm{w}}$ is a convex combination of $\bm{w}(k)$ and $\bm{w}(k+1)$, thus the $2$-norm of each component of it is bounded by $B$), we have
\begin{align*}
\left\|\frac{\partial s_p(\bm{w})}{\partial \bm{w}_r}-\frac{\partial s_p(\hat{\bm{w}})}{\partial \bm{w}_r} \right\|_2 &\lesssim \frac{1}{\sqrt{n_1m}}(B^2+B+1) \|\bm{w}_r-\hat{\bm{w}}_r\|_2\lesssim \frac{B^2}{\sqrt{n_1m}}\|\bm{w}_r-\hat{\bm{w}}_r\|_2.
\end{align*}
With this result, we can deduce that
\begin{equation}
\begin{aligned}
|\bm{I}_1^p(k+1)|&= \left|\sum\limits_{r=1}^m \int_{0}^{\eta} \left\langle \frac{\partial L(\bm{w}(k+1))}{\partial \bm{w}_r}, \frac{\partial s_p(\bm{w}(k+1)+\alpha\frac{\partial L(\bm{w}(k+1))}{\partial \bm{w}} )}{\partial \bm{w}_r} -\frac{\partial s_p(\bm{w}(k+1))}{\partial \bm{w}_r} \right\rangle d\alpha\right|\\
&\leq \sum\limits_{r=1}^m \int_{0}^{\eta} \left\|\frac{\partial L(\bm{w}(k+1))}{\partial \bm{w}_r}\right\|_2 \left\|\frac{\partial s_p(\bm{w}(k+1)+\alpha\frac{\partial L(\bm{w}(k+1))}{\partial \bm{w}} )}{\partial \bm{w}_r} -\frac{\partial s_p(\bm{w}(k+1))}{\partial \bm{w}_r}\right\|_2d\alpha\\
&\lesssim\frac{B^2}{\sqrt{n_1m}}\sum\limits_{r=1}^m \int_{0}^{\eta} \left\|\frac{\partial L(\bm{w}(k+1))}{\partial \bm{w}_r}\right\|_2 |\alpha|\left\|\frac{\partial L(\bm{w}(k+1))}{\partial \bm{w}_r}\right\|_2d\alpha\\
&\lesssim \frac{\eta^2 B^2}{\sqrt{n_1m}}\sum\limits_{r=1}^m  \left\|\frac{\partial L(\bm{w}(k+1))}{\partial \bm{w}_r}\right\|_2^2 .
\end{aligned}
\end{equation}
Recall that
\[\frac{\partial L(\bm{w}(k+1))}{\partial \bm{w}_r} =\sum\limits_{p=1}^{n_1} s_p(k+1)\frac{\partial s_p(k+1)}{\partial \bm{w}_r}+\sum\limits_{j=1}^{n_2} h_j(k+1)\frac{\partial h_j(k+1)}{\partial \bm{w}_r}.\]
From the estimation of $\|\bm{w}_r(t+1)-\bm{w}_r(t)\|_2$ in the proof of Lemma 3.7, we have that
\begin{equation}
\left\|\frac{\partial L(\bm{w}(k+1))}{\partial \bm{w}_r}\right\|_2\lesssim \frac{B^2}{\sqrt{m}} \sqrt{L(k+1)}.	
\end{equation}
Therefore, plugging (6.20) into (6.19), we have
\begin{equation}
\begin{aligned}
\|\bm{I}_1(k+1)\|_2& =\sqrt{ \sum\limits_{p=1}^{n_1}|\bm{I}_1^p(k+1)|^2}\\
&\lesssim  \frac{\eta^2B^2}{\sqrt{m}}\sum\limits_{r=1}^m \frac{B^4}{m} L(k+1)\\
&\lesssim \frac{\eta^2  B^6}{\sqrt{m}} L(k+1).
\end{aligned}
\end{equation}
Similarly, we can deduce that
\begin{align*}
\left\|\frac{\partial h_j(\bm{w})}{\partial \bm{w}_r}-\frac{\partial h_j(\hat{\bm{w}})}{\partial \bm{w}_r} \right\|_2 &\lesssim \frac{1}{\sqrt{n_2m}} \|\bm{w}_r-\hat{\bm{w}}_r\|_2.
\end{align*}
Thus
\begin{equation}
\|\bm{I}_2(k+1)\|_2 \lesssim \frac{\eta^2 B^4 }{\sqrt{m}} L(k+1).
\end{equation}
\end{proof}

\subsection{Proof of Theorem 3.1}
\begin{proof}
We first recall the recursive formula of the implicit gradient descent:
\[\begin{bmatrix}
\bm{s}(\bm{w}(k+1))\\ \bm{h}(\bm{w}(k+1))
\end{bmatrix}
=(\bm{I}+\eta \bm{G}(k+1))^{-1}\left( \begin{bmatrix}
\bm{s}(\bm{w}(k))\\ \bm{h}(\bm{w}(k))
\end{bmatrix} - \begin{bmatrix}
\bm{I}_1(k+1)\\ \bm{I}_2(k+1)
\end{bmatrix}\right).\]
Thus,
\begin{equation}
\begin{aligned}
\left\|\begin{bmatrix}
    \bm{s}(\bm{w}(k+1))\\ \bm{h}(\bm{w}(k+1))
\end{bmatrix} \right\|_2^2  &= \left\|(\bm{I}+\eta \bm{G}(k+1))^{-1}\left( \begin{bmatrix}
    \bm{s}(\bm{w}(k))\\ \bm{h}(\bm{w}(k))
\end{bmatrix} - \begin{bmatrix}
    \bm{I}_1(k+1)\\ \bm{I}_2(k+1)
\end{bmatrix}\right)\right\|_2^2 \\
&\leq \frac{1}{\left(1+\frac{\eta \lambda_0}{2}\right)^2} \left[ \left\| \begin{bmatrix}
    \bm{s}(\bm{w}(k))\\ \bm{h}(\bm{w}(k))
\end{bmatrix}\right\|_2^2 -2\left\langle \begin{bmatrix}
    \bm{s}(\bm{w}(k))\\ \bm{h}(\bm{w}(k))
\end{bmatrix}, \begin{bmatrix}
    \bm{I}_1(k+1)\\ \bm{I}_2(k+1)
\end{bmatrix} \right\rangle + \left\| \begin{bmatrix}
    \bm{I}_1(k+1)\\ \bm{I}_2(k+1)
\end{bmatrix}\right\|_2^2 \right] \\
&\leq \frac{1}{\left(1+\frac{\eta \lambda_0}{2}\right)^2} \left[ \left\| \begin{bmatrix}
    \bm{s}(\bm{w}(k))\\ \bm{h}(\bm{w}(k))
\end{bmatrix}\right\|_2^2 +2\left\| \begin{bmatrix}
    \bm{s}(\bm{w}(k))\\ \bm{h}(\bm{w}(k))
\end{bmatrix}\right\|_2 \left\| \begin{bmatrix}
    \bm{I}_1(k+1)\\ \bm{I}_2(k+1)
\end{bmatrix} \right\|_2 + \left\| \begin{bmatrix}
    \bm{I}_1(k+1)\\ \bm{I}_2(k+1)
\end{bmatrix}\right\|_2^2 \right].
\end{aligned}
\end{equation}

From the proof of Lemma 3.9, combining (6.21) and (6.22), we can deduce that
\begin{equation}
\left\| \begin{bmatrix} \bm{I}_1(k+1)\\ \bm{I}_2(k+1) \end{bmatrix} \right\|_2\leq \frac{C\eta^2 B^6}{\sqrt{m}} L(k+1)\leq  \frac{C\eta^2  B^6}{\sqrt{m}} \sqrt{L(0)L(k+1)}
\end{equation}
and
\begin{equation}
\left\| \begin{bmatrix} \bm{I}_1(k+1)\\  \bm{I}_2(k+1) \end{bmatrix} \right\|_2^2 \leq \frac{C^2\eta^4 B^{12}}{m} L(k+1)^2\leq \frac{C^2\eta^2  B^{12}}{\sqrt{m}} L(0)L(k+1).
\end{equation}
Plugging (6.24) and (6.25) into (6.23) yields that
\begin{equation}
\begin{aligned}
L(k+1)&\leq \frac{1}{\left(1+\frac{\eta\lambda_0}{2}\right)^2} \left[L(k)+2\sqrt{L(k)}\left\| \begin{bmatrix}
    \bm{I}_1(k+1)\\ \bm{I}_2(k+1)
\end{bmatrix}\right\|_2   + \left\| \begin{bmatrix}
    \bm{I}_1(k+1)\\ \bm{I}_2(k+1)
\end{bmatrix}\right\|_2^2\right] \\
&\leq \frac{1}{\left(1+\frac{\eta\lambda_0}{2}\right)^2} \left[ L(k)+\frac{2C\eta^2  B^6}{\sqrt{m}} \sqrt{L(k)}\sqrt{L(0)}\sqrt{L(k+1)}+\frac{C^2\eta^2  B^{12}}{\sqrt{m}}L(0) L(k+1). \right]\\
&\leq \frac{1}{\left(1+\frac{\eta\lambda_0}{2}\right)^2} \left[ L(k)+\frac{C\eta^2  B^6}{\sqrt{m}} \sqrt{L(0)}(L(k)+L(k+1))+\frac{C^2\eta^2 B^{12}}{\sqrt{m}}L(0) L(k+1), \right]
\end{aligned}
\end{equation}
where the second inequality follows from the mean inequality $2\sqrt{ab}\leq a+b$ for $a,b\geq 0$.

Let $f(m)=\frac{C\eta^2  B^6}{\sqrt{m}}\sqrt{L(0)}$, a simple algebraic transformations for (6.26) leads to that
\begin{align*}
L(k+1)&\leq \frac{1+f(m)}{\left(1+\frac{\eta\lambda_0}{2}\right)^2-f(m)-f^2(m)} L(k) \\
&\leq \frac{1+\frac{\eta\lambda_0}{4}}{\left(1+\frac{\eta\lambda_0}{2}\right)^2-\frac{\eta\lambda_0}{4}-\frac{\eta^2\lambda_0^2}{16}} L(k) \\
&\leq \frac{1+\frac{\eta\lambda_0}{4}}{1+\frac{3\eta\lambda_0}{4}+\frac{3\eta^2\lambda_0^2}{16}} L(k)\\
&\leq \frac{1+\frac{\eta\lambda_0}{4}}{1+\frac{3\eta\lambda_0}{4}+\frac{\eta^2\lambda_0^2}{8}} L(k)\\
&=\frac{1+\frac{\eta\lambda_0}{4}}{\left(1+\frac{\eta\lambda_0}{4}\right)\left(1+\frac{\eta\lambda_0}{2}\right)} L(k)\\
&= \left(1+\frac{\eta\lambda_0}{2}\right)^{-1}L(k) \\
&\leq \left(1+\frac{\eta\lambda_0}{2}\right)^{-(k+1)}L(0),
\end{align*}
where the second inequality is due to that we choose $m$ large enough such that $f(m)\leq \frac{\eta \lambda_0}{4}$, i.e.,
\[\frac{C\eta^2 B^6}{\sqrt{m}}\sqrt{L(0)} \leq \frac{\eta \lambda_0}{4}.\]
From the estimation for $L(0)$ in Lemma 3.6, we have
\begin{equation}
m=\Omega\left(\frac{\eta^2B^{12}L(0)}{\lambda_0^2}\right)=\Omega\left(\frac{\eta^2d^{8} \left(\log\left(\frac{m}{\delta}\right)\right)^6\log\left(\frac{n_1+n_2}{\delta}\right) }{\lambda_0^2}\right).
\end{equation}
Combining the requirements for $m$, i.e., (6.18) and (6.27), we can deduce that
\[m=\Omega\left(\frac{d^8}{\lambda_0^2}\log\left(\frac{n_1+n_2}{\delta}\right)\max\left\{\eta^4 d^2 \log\left(\frac{n_1+n_2}{\delta}\right), \frac{1}{\lambda_0^2} \left(\log\left(\frac{m}{\delta}\right)\right)^2, \frac{\eta^2 \left(\log\left(\frac{m}{\delta}\right)\right)^6\log\left(\frac{n_1+n_2}{\delta}\right) }{\lambda_0^2}  \right\}  \right) . \]	
\end{proof}


\begin{thebibliography}{99}

\bibitem{10}
K.~He, X.~Zhang, S.~Ren, and J.~Sun, ``Deep residual learning for image
  recognition,'' in \emph{Proceedings of the IEEE conference on Computer Vision
  and pattern recognition}, 2016, pp. 770--778.

\bibitem{11}
J.~Devlin, M.-W. Chang, K.~Lee, and K.~Toutanova, ``BERT: Pre-training of deep
  bidirectional transformers for language understanding,'' \emph{arXiv preprint
  arXiv:1810.04805}, 2018.

\bibitem{12}
D.~Silver, A.~Huang, C.~J. Maddison, A.~Guez, L.~Sifre, G.~Van Den~Driessche,
  J.~Schrittwieser, I.~Antonoglou, V.~Panneershelvam, M.~Lanctot \emph{et~al.},
  ``Mastering the game of go with deep neural networks and tree search,''
  \emph{Nature}, vol. 529, no. 7587, pp. 484--489, 2016.

\bibitem{6}
M.~Raissi, P.~Perdikaris, and G.~E. Karniadakis, ``Physics-informed neural
  networks: A deep learning framework for solving forward and inverse problems
  involving nonlinear partial differential equations,'' \emph{Journal of
  Computational Physics}, vol. 378, pp. 686--707, 2019.

\bibitem{26}
B.~Yu \emph{et~al.}, ``The deep Ritz method: a deep learning-based numerical
  algorithm for solving variational problems,'' \emph{Communications in
  Mathematics and Statistics}, vol.~6, no.~1, pp. 1--12, 2018.

\bibitem{27}
Y.~Zang, G.~Bao, X.~Ye, and H.~Zhou, ``Weak adversarial networks for
  high-dimensional partial differential equations,'' \emph{Journal of
  Computational Physics}, vol. 411, p. 109409, 2020.

\bibitem{28}
J.~W. Siegel, Q.~Hong, X.~Jin, W.~Hao, and J.~Xu, ``Greedy training algorithms
  for neural networks and applications to pdes,'' \emph{Journal of
  Computational Physics}, vol. 484, p. 112084, 2023.

\bibitem{25}
I.~E. Lagaris, A.~Likas, and D.~I. Fotiadis, ``Artificial neural networks for
  solving ordinary and partial differential equations,'' \emph{IEEE
  Transactions on Neural Networks}, vol.~9, no.~5, pp. 987--1000, 1998.

\bibitem{13}
J.~Han, A.~Jentzen, and W.~E, ``Solving high-dimensional partial differential
  equations using deep learning,'' \emph{Proceedings of the National Academy of
  Sciences}, vol. 115, no.~34, pp. 8505--8510, 2018.

\bibitem{17}
D.~Lucor, A.~Agrawal, and A.~Sergent, ``Simple computational strategies for
  more effective physics-informed neural networks modeling of turbulent natural
  convection,'' \emph{Journal of Computational Physics}, vol. 456, p. 111022,
  2022.

\bibitem{14}
D.~Gilton, G.~Ongie, and R.~Willett, ``Neumann networks for linear inverse
  problems in imaging,'' \emph{IEEE Transactions on Computational Imaging},
  vol.~6, pp. 328--343, 2019.

\bibitem{15}
Y.~Fan and L.~Ying, ``Solving electrical impedance tomography with deep
  learning,'' \emph{Journal of Computational Physics}, vol. 404, p. 109119,
  2020.

\bibitem{16}
L.~Lu, R.~Pestourie, W.~Yao, Z.~Wang, F.~Verdugo, and S.~G. Johnson,
  ``Physics-informed neural networks with hard constraints for inverse
  design,'' \emph{SIAM Journal on Scientific Computing}, vol.~43, no.~6, pp.
  B1105--B1132, 2021.

\bibitem{18}
M.~Germain, H.~Pham, X.~Warin \emph{et~al.}, ``Neural networks-based algorithms
  for stochastic control and pdes in finance,'' \emph{arXiv preprint
  arXiv:2101.08068}, 2021.

\bibitem{19}
J.~B. Heaton, N.~G. Polson, and J.~H. Witte, ``Deep learning for finance: deep
  portfolios,'' \emph{Applied Stochastic Models in Business and Industry},
  vol.~33, no.~1, pp. 3--12, 2017.

\bibitem{20}
Y.~Bai, T.~Chaolu, and S.~Bilige, ``The application of improved
  physics-informed neural network (ipinn) method in finance,'' \emph{Nonlinear
  Dynamics}, vol. 107, no.~4, pp. 3655--3667, 2022.

\bibitem{21}
J.~Sirignano and K.~Spiliopoulos, ``Dgm: A deep learning algorithm for solving
  partial differential equations,'' \emph{Journal of Computational Physics},
  vol. 375, pp. 1339--1364, 2018.

\bibitem{2}
S.~S. Du, X.~Zhai, B.~Poczos, and A.~Singh, ``Gradient descent provably
  optimizes over-parameterized neural networks,'' \emph{arXiv preprint
  arXiv:1810.02054}, 2018.

\bibitem{3}
S.~Du, J.~Lee, H.~Li, L.~Wang, and X.~Zhai, ``Gradient descent finds global
  minima of deep neural networks,'' in \emph{International Conference on
  Machine Learning}.\hskip 1em plus 0.5em minus 0.4em\relax PMLR, 2019, pp.
  1675--1685.

\bibitem{4}
Y.~Cao and Q.~Gu, ``Generalization error bounds of gradient descent for
  learning over-parameterized deep RELU networks,'' in \emph{Proceedings of the
  AAAI Conference on Artificial Intelligence}, vol.~34, no.~04, 2020, pp.
  3349--3356.

\bibitem{29}
D.~Zou and Q.~Gu, ``An improved analysis of training over-parameterized deep
  neural networks,'' \emph{Advances in Neural Information Processing Systems},
  vol.~32, 2019.

\bibitem{30}
Y.~Cao and Q.~Gu, ``Generalization bounds of stochastic gradient descent for
  wide and deep neural networks,'' \emph{Advances in Neural Information
  Processing Systems}, vol.~32, 2019.

\bibitem{1}
A.~Jacot, F.~Gabriel, and C.~Hongler, ``Neural tangent kernel: Convergence and
  generalization in neural networks,'' \emph{Advances in Neural Information
  Processing Systems}, vol.~31, 2018.

\bibitem{9}
Y.~Gao, Y.~Gu, and M.~Ng, ``Gradient descent finds the global optima of
  two-layer physics-informed neural networks,'' in \emph{International
  Conference on Machine Learning}. PMLR,
  2023, pp. 10\,676--10\,707.

\bibitem{23}
Y.~Li, S.-C. Chen, and S.-J. Huang, ``Implicit stochastic gradient descent for
  training physics-informed neural networks,'' in \emph{Proceedings of the AAAI
  Conference on Artificial Intelligence}, vol.~37, no.~7, 2023, pp. 8692--8700.

\bibitem{7}
J.~M{\"u}ller and M.~Zeinhofer, ``Achieving high accuracy with pinns via energy
  natural gradient descent,'' in \emph{International Conference on Machine
  Learning}. PMLR, 2023, pp.
  25\,471--25\,485.

\bibitem{8}
T.~Luo and H.~Yang, ``Two-layer neural networks for partial differential
  equations: Optimization and generalization theory,'' \emph{arXiv preprint
  arXiv:2006.15733}, 2020.

\bibitem{33}
A.~D. Jagtap, K.~Kawaguchi, and G.~E. Karniadakis, ``Adaptive activation
  functions accelerate convergence in deep and physics-informed neural
  networks,'' \emph{Journal of Computational Physics}, vol. 404, p. 109136,
  2020.

\bibitem{34}
T.~De~Ryck, S.~Lanthaler, and S.~Mishra, ``On the approximation of functions by
  tanh neural networks,'' \emph{Neural Networks}, vol. 143, pp. 732--750, 2021.

\bibitem{35}
I.~G{\"u}hring and M.~Raslan, ``Approximation rates for neural networks with
  encodable weights in smoothness spaces,'' \emph{Neural Networks}, vol. 134,
  pp. 107--130, 2021.

\bibitem{37}
X.~Xu, T.~Du, W.~Kong, Y.~Li, and Z.~Huang, ``Convergence analysis of natural
  gradient descent for over-parameterized physics-informed neural networks,''
  \emph{arXiv preprint arXiv:2408.00573}, 2024.

\bibitem{32}
Z.~Song and X.~Yang, ``Quadratic suffices for over-parametrization via matrix
  chernoff bound,'' \emph{arXiv preprint arXiv:1906.03593}, 2019.

\bibitem{36}
S.~Arora, S.~Du, W.~Hu, Z.~Li, and R.~Wang, ``Fine-grained analysis of
  optimization and generalization for overparameterized two-layer neural
  networks,'' in \emph{International Conference on Machine Learning}. PMLR, 2019, pp. 322--332.

\bibitem{24}
A.~K. Kuchibhotla and A.~Chakrabortty, ``Moving beyond sub-gaussianity in
  high-dimensional statistics: Applications in covariance estimation and linear
  regression,'' \emph{Information and Inference: A Journal of the IMA},
  vol.~11, no.~4, pp. 1389--1456, 2022.

\end{thebibliography}
\end{document}